\pdfoutput=1

\documentclass[11pt,a4paper]{article}

\usepackage{times,latexsym}
\usepackage{url}

\usepackage[acceptedWithA]{tacl2021v1}
\usepackage[T1]{fontenc}

\usepackage{times}
\usepackage{latexsym}
\usepackage{bm}

\usepackage[inline]{enumitem}

\usepackage[T1]{fontenc}

\usepackage[utf8]{inputenc}
\usepackage{microtype}
\usepackage{graphicx}
\usepackage{booktabs}
\usepackage{amssymb}
\usepackage{amsmath}
\usepackage{multirow}
\usepackage{booktabs}
\usepackage{subfigure}
\usepackage{tikz}
\usetikzlibrary{positioning, patterns, patterns.meta, arrows.meta, shapes.geometric,arrows,positioning,calc,fit}
\usepackage{listings}
\usepackage{xcolor, colortbl}
\usepackage{array}
\usepackage{etoolbox}
\usepackage{xspace}

\usepackage{pifont}
\newcommand{\cmark}{\ding{51}}%
\newcommand{\xmark}{\ding{55}}%

\newcommand{\mynatop}[1]{%
  \ifstrequal{#1}{alternation}{\mathrel{\downharpoonleft\!\upharpoonright}}{%
  \ifstrequal{#1}{equiv}{\equiv}{%
  \ifstrequal{#1}{forwardentailment}{\sqsubseteq}{%
  \ifstrequal{#1}{reventailment}{\sqsupseteq}{%
  \ifstrequal{#1}{negation}{\curlywedge}{%
  \ifstrequal{#1}{independence}{\#}{%
  \ifstrequal{#1}{cover}{\smile}{%
  \PackageError{mynatop}{Invalid option: #1}{Valid options are: alternation, equiv, forwardentailment, reventailment, negation, independence, cover}%
  }}}}}}}%
}

\usepackage{xspace,mfirstuc,tabulary}

\newif\iftaclinstructions
\taclinstructionsfalse 
\iftaclinstructions
\renewcommand{\confidential}{}
\renewcommand{\anonsubtext}{(No author info supplied here, for consistency with
TACL-submission anonymization requirements)}
\newcommand{\instr}
\fi

\iftaclpubformat 

\else

\fi

\author{Rami Aly\\
  University of Cambridge \\
 Department of Computer Science\\
 and Technology \\
  \texttt{rami.aly@cl.cam.ac.uk} \\\And
  Andreas Vlachos \\
University of Cambridge \\
 Department of Computer Science\\
 and Technology\\
  \texttt{andreas.vlachos@cl.cam.ac.uk}
}

\definecolor{listingbg}{rgb}{0.97,0.97,0.97}
\definecolor{listingexample}{rgb}{0.55,0.7,0.9}
\definecolor{codegreen}{rgb}{0,0.6,0}
\definecolor{codegray}{rgb}{0.5,0.5,0.5}
\definecolor{codepurple}{rgb}{0.58,0,0.82}
\lstdefinestyle{mystyle}{
    backgroundcolor=\color{listingbg},
    commentstyle=\color{codegreen},
    keywordstyle=\color{magenta},
    numberstyle=\tiny\color{codegray},
    stringstyle=\color{codepurple},
    basicstyle=\fontsize{8pt}{9pt}\sffamily,
    breakatwhitespace=false,
    breaklines=true,
    captionpos=b,
    keepspaces=true,
    numbers=left,
    numbersep=7pt,
    showspaces=false,
    showstringspaces=false,
    showtabs=false,
    tabsize=2,
    breakindent=0pt,
    frame=lines,
    aboveskip=\baselineskip,
    framesep=10pt,
    captionpos=b,
    abovecaptionskip=10pt,
    moredelim=**[is][\color{listingexample}]{@}{@},
    columns=fullflexible,
    xleftmargin=1cm,
}
\newcommand{\coloredcircle}[2]{%
\raisebox{-0.3em}{
  \begin{tikzpicture}
    \fill [#1] (0, 0) circle (0.2cm);
    \node at (0,0) {#2};
  \end{tikzpicture}%
  }
}

\usepackage{inconsolata}

\definecolor{celadon}{rgb}{0.67, 0.88, 0.69}
\definecolor{amethyst}{rgb}{1.0, 0.49, 0.0} 
\definecolor{bondiblue}{rgb}{0.0, 0.68, 0.71} 
\definecolor{aquamarine}{rgb}{0.13, 0.61, 0.94}
\definecolor{babypink}{rgb}{0.96, 0.76, 0.76}
\definecolor{bittersweet}{rgb}{1.0, 0.44, 0.37}
\definecolor{paperblue}{HTML}{006666}

\definecolor{brightgreen}{rgb}{0.4, 1.0, 0.0}
\definecolor{celadon}{rgb}{0.67, 0.82, 0.60}

\newcommand{\projectname}{\mbox{\textsc{TabVer}}\xspace}

\newcommand{\feverousaccfull}{$71.4$}
\newcommand{\feverousaccleadfull}{$3.4$}
\newcommand{\feverousaccleadfullSymbol}{$10.5$}
\newcommand{\feverousfleadfull}{$5.6$}
\newcommand{\feverousaccleadnumerical}{$5.6$}

\newcommand{\tabfactaccleadfull}{$0.5$}
\newcommand{\tabfactfleadfull}{$0.3$} 

\title{\projectname: Tabular Fact Verification with Natural Logic}

\begin{document}

\maketitle

\begin{abstract}

Fact verification on tabular evidence incentivises the use of symbolic reasoning models where a logical form is constructed (e.g.\ a LISP-style program), providing greater verifiability than fully neural approaches. However, these systems typically rely on well-formed tables, restricting their use in many scenarios.
An emerging symbolic reasoning paradigm for textual evidence focuses on natural logic inference, which constructs proofs by modelling set-theoretic relations between a claim and its evidence in natural language. This approach provides flexibility and transparency but is less compatible with tabular evidence since the relations do not extend to arithmetic functions.
We propose a set-theoretic interpretation of numerals and arithmetic functions in the context of natural logic, enabling the integration of arithmetic expressions in deterministic proofs. We leverage large language models to generate arithmetic expressions by generating questions about salient parts of a claim which are answered by executing appropriate functions on tables. In a few-shot setting on FEVEROUS, we achieve an accuracy of \feverousaccfull, outperforming both fully neural and symbolic reasoning models
by \feverousaccleadfull \ points. When evaluated on TabFact
without any further training, our method remains competitive with an accuracy lead of \tabfactaccleadfull \ points.

\end{abstract}

\section{Introduction}
\label{sec:intro}

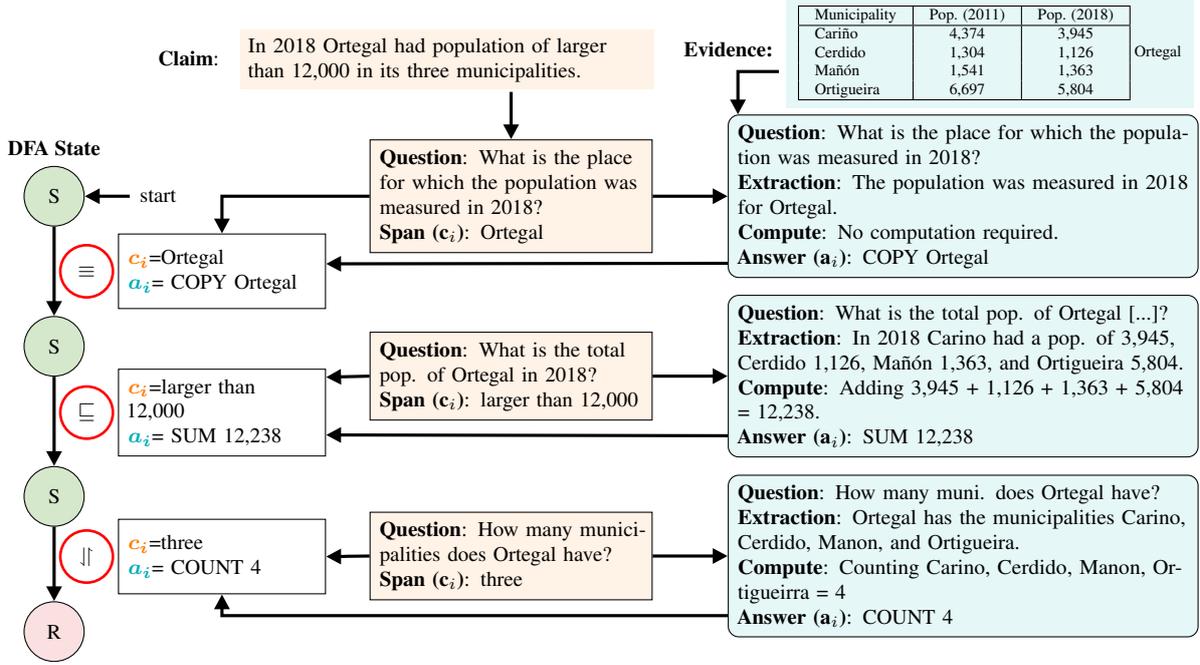
\begin{figure*}[ht!]
\resizebox{1\linewidth}{!}{
\begin{tikzpicture}[every node/.style={font=\scriptsize}, node distance=1.2cm and 2cm, 
    s/.style={circle, draw, minimum size=0.8cm},
    arithexp/.style={rectangle, draw, minimum height=1cm, text width=2.5cm, align=left},
    qa/.style={draw, rounded corners, align=center, text width=6cm, align=left, fill=bondiblue!10, 
    font=\scriptsize},
    question/.style={rectangle, text width=3.5cm, draw, align=left, fill=amethyst!10, 
    align=left},
    natop/.style={circle, draw=red, line width=1pt,minimum size=0.7cm,inner sep=0pt},]

    \node[s, fill=celadon!50] (s0) {S};
    \node[s, below=1.2cm of s0, fill=celadon!50] (s1) {S};
    \node[s, below=1.2cm of s1, fill=celadon!50] (s2) {S};
    \node[draw, circle, minimum size=0.8cm, below=1.0cm of s2, fill=babypink!50] (reject) {R};
    \node[right=0.6cm of s0] (start) {start};

    \node[question, anchor=west, shift={(4.2cm, 0cm)}] at (s0) (q1) {\textbf{Question}: What is the place for which the population was measured in 2018?\\ \textbf{Span (c$_i$)}: Ortegal};
    \node[question, anchor=west, shift={(4.2cm, -0.4cm)}] at (s1) (q2) {\textbf{Question}: What is the total pop. of Ortegal in 2018?\\ \textbf{Span (c$_i$)}: larger than 12,000};
    \node[question, anchor=west, shift={(4.2cm, -0.8cm)}] at (s2) (q3) {\textbf{Question}: How many municipalities does Ortegal have?\\ \textbf{Span (c$_i$)}: three};

    \node[qa, right=1cm of q1] (ex1) {\textbf{Question}: What is the place for which the population was measured in 2018?\\\textbf{Extraction}: The population was measured in 2018 for Ortegal.\\ \textbf{Compute}: No computation required.\\ \textbf{Answer (a$_i$)}: COPY Ortegal};
    \node[qa, right=1cm of q2] (ex2) {\textbf{Question}: What is the total pop. of Ortegal [...]? \\ \textbf{Extraction}: In 2018 Carino had a pop. of 3,945, Cerdido 1,126, Mañón 1,363, and Ortigueira 5,804.\\ \textbf{Compute}: Adding 3,945 + 1,126 + 1,363 + 5,804 = 12,238.\\ \textbf{Answer (a$_i$)}: SUM 12,238};
    \node[qa, right=1cm of q3] (ex3) {\textbf{Question}: How many muni. does Ortegal have? \\ \textbf{Extraction}: Ortegal has the municipalities Carino, Cerdido, Manon, and Ortigueira.\\ \textbf{Compute}: Counting Carino, Cerdido, Manon, Ortigueirra = 4\\ \textbf{Answer (a$_i$)}: COUNT 4};

    \node[arithexp, anchor=north west, shift={(0.85cm,-0.5cm)}] at (s0) (a1) {\textcolor{amethyst}{\bm{$c_i$}}=Ortegal \\ \textcolor{bondiblue}{\bm{$a_i$}}= COPY Ortegal};
    \node[arithexp, anchor=north west, shift={(0.85cm,-0.3cm)}] at (s1) (a2) {\textcolor{amethyst}{\bm{$c_i$}}=larger than 12,000 \\ \textcolor{bondiblue}{\bm{$a_i$}}= SUM 12,238};
    \node[arithexp, anchor=north west, shift={(0.85cm,-0.3cm)}] at (s2) (a3) {\textcolor{amethyst}{\bm{$c_i$}}=three \\ \textcolor{bondiblue}{\bm{$a_i$}}= COUNT 4};

    \node[natop, left=0.05cm of a1] (natop1) {$\mynatop{equiv}$};
    \node[natop, left=0.05cm of a2] (natop2) {$\mynatop{forwardentailment}$};
    \node[natop, left=0.05cm of a3] (natop3) {$\mynatop{alternation}$};

    \node[above =0cm of s0, anchor=south] (label) {\textbf{DFA State}};
     \node[ text width=8cm, anchor=south, shift={(-0.7cm,1.3cm)}] at (q1) (claim) {\textbf{Claim}: { \colorbox{amethyst!10}{\parbox{5.2cm}{In 2018 Ortegal had population of larger than 12,000 in its three municipalities.} }
     }
     };
     \node [anchor=north, shift={(-0.3cm,2.8cm)}] at (ex1) (evidence)  {
        \textbf{Evidence:} { \tiny
        \colorbox{bondiblue!10}{
        \begin{tabular}{|l|c|c|}
        \hline
        Municipality & Pop. (2011) & Pop. (2018) \\
        \hline
        Cariño &  4,374 & 3,945\\
        Cerdido & 1,304 & 1,126 \\
        Mañón  & 1,541 & 1,363 \\
        Ortigueira  & 6,697 & 5,804 \\
        \hline
        \end{tabular}
        Ortegal
        }
        }
    };

    \draw[-{Triangle[length=2mm, width=2mm]}, very thick] (start) -- (s0);
    \draw[-{Triangle[length=2mm, width=2mm]}, very thick] (s0) -- (s1);
    \draw[-{Triangle[length=2mm, width=2mm]}, very thick] (s1) -- (s2);
    \draw[-{Triangle[length=2mm, width=2mm]}, very thick] (s2) -- (reject);

    \draw[-{Triangle[length=2mm, width=2mm]}, very thick] (claim.south) + (0.7, +0.1) -- (q1.north);
    \draw[-{Triangle[length=2mm, width=2mm]}, very thick] ([xshift=+1.1cm, yshift=-0.45cm]evidence.west) + (0.3, 0.2) -| ([xshift=-3cm, yshift=0cm]ex1.north);
    
    \draw[-{Triangle[length=2mm, width=2mm]}, very thick] (q1.west) -| (a1.north);
    \draw[-{Triangle[length=2mm, width=2mm]}, very thick] (q2.west) to[out=180, in=30, looseness=0.0] ([xshift=0cm, yshift=0.475cm]a2.east);
    \draw[-{Triangle[length=2mm, width=2mm]}, very thick] (q3.west) to[out=180, in=0, looseness=0.1] (a3.east);

    \draw[-{Triangle[length=2mm, width=2mm]}, very thick] (q1.east) -- (ex1.west);
    \draw[-{Triangle[length=2mm, width=2mm]}, very thick] (q2.east) -- (ex2.west);
    \draw[-{Triangle[length=2mm, width=2mm]}, very thick] (q3.east) -- (ex3.west);

    \draw[-{Triangle[length=2mm, width=2mm]}, very thick] ([xshift=-0cm, yshift=-0.9cm]ex1.west) to[out=-45, in=0, looseness=0] ([xshift=0cm, yshift=0.1cm]a1.east);
    \draw[-{Triangle[length=2mm, width=2mm]}, very thick] ([xshift=0cm, yshift=-0.8cm]ex2.west) to[out=-45, in=330, looseness=0] ([xshift=0cm, yshift=-0.3cm]a2.east);
    \draw[-{Triangle[length=2mm, width=2mm]}, very thick] ([xshift=0cm, yshift=-0.8cm]ex3.west) -| (a3.south);

\end{tikzpicture}}
\vspace{-0.9em}
\caption{ High-level illustration of \projectname. \projectname \ proposes a set-theoretic view on numerals and arithmetic functions, which is integrated into natural logic proofs as arithmetic comparisons between claim and answers to questions (ArithExps), resulting in deterministic inference (left). To generate ArithExps, \projectname \ asks questions about salient parts $c_i$ of a claim (\textcolor{amethyst}{middle}). The questions are answered using tabular evidence $E$, by generating a rationale and a set-theoretic compatible representation of required computations (\textcolor{bondiblue}{right}).
}
\label{fig:method-overview}
\vspace{-0.5em}
\end{figure*}

Fact verification systems assess the veracity of claims based on evidence and provide an explanation for the prediction. In the case of tabular evidence, verification frequently relies on symbolic reasoning steps, such as the execution of arithmetic functions, to accurately predict whether a claim is supported by evidence \citep[][\textit{inter alia}]{herzig-etal-2020-tapas}. 
This incentivises symbolic reasoning systems, where a logical representation of a claim and its tabular evidence (e.g.\ a LISP-style program) is executed to produce the veracity prediction \citep{chen-etal-2020-tabfact, cheng-etal-2023-binding}. Since the execution of these logical forms is deterministic, they serve as faithful explanations of the model's reasoning \citep{jacovi-goldberg-2021-aligning}.
However, these systems typically rely on well-formed tables, constraining their use in many scenarios, such as reasoning over diverse tabular structures as typically found on Wikipedia.
Consequently, the majority of recently proposed verification models focus on neural entailment models that latently execute arithmetic functions \citep{liu-etal-2022-tapex, gu-etal-2022-pasta} 
or generate a natural language explanation alongside its prediction \citep[][\textit{inter alia}]{wei-etal-2022-chainofthought}. While systems that produce natural language explanations
are more flexible regarding the evidence format, they do not necessarily generate faithful explanations \citep{atanasova-etal-2023-faithfulness}.%

An emergent symbolic reasoning paradigm for textual evidence focuses on logical inference by directly comparing claim and textual evidence via natural logic inference \citep{angeli-manning-2014-naturalli}, achieving high prediction accuracy while maintaining faithful explanations \citep{krishna-etal-2022-proofver, aly-etal-2023-qanatver, strong-etal-2024-zeroshot}.
However, current natural logic systems are unable to handle tabular evidence since the semantic relationship captured between aligned claim-evidence spans via natural logic's 
set-theoretic operators does not extend to arithmetic functions \citep{maccartney-manning-2009-extended}. For instance, in Figure~\ref{fig:method-overview}, no evidence in the table directly corresponds to the part of the claim that states \emph{three municipalities}. Instead, arithmetic computation on the table, beyond the expressiveness of natural logic's set-theoretic operators, is required (i.e.\ counting relevant cells in this example).

To this end, we propose \projectname: %
\textbf{Tab}ular Fact \textbf{Ver}ification, a natural logic inference system that adds arithmetic reasoning capabilities to reason over tabular evidence directly in natural language. We define a set-theoretic interpretation of comparisons between numerals in claim-evidence pairs, and extend that definition to executions of arithmetic functions via arithmetic expressions (ArithExps) to enable their integration into natural logic proofs. The proofs are executed deterministically on a finite state automaton (DFA) as defined in natural logic inference.
ArithExps are produced by leveraging large language models \citep[][\textit{inter alia}]{brown-etal-2020-language}, 
generating questions about salient parts of the claim $c_i$, which are answered via a rationale %
that produces an answer $a_i$. %
As illustrated in Figure~\ref{fig:method-overview}, \projectname \ will generate a question such as ``\emph{What is the total population of Ortegal in 2018}'' to verify the part \emph{larger than 12000} in the claim $c$. Answering this question on the evidence table produces a rationale with the expression $\text{SUM } \emph{12,238}$ as the final answer $a_i$, indicating the execution of the function $\text{SUM}(3945, 1126, 1363, 5804) = 12238$ over relevant evidence in $E$. 
The aligned pair (\emph{larger than 12000}, $\emph{SUM } 12,238$) is then assigned a natural logic operator as part of a natural logic proof, with the predicted operator being consistent with our set-theoretic definitions (Figure~\ref{fig:set-theoretic-definition}).

In a few-shot setting with $64$ training instances
on the tabular subset of the FEVEROUS dataset \citep{aly-etal-2021-feverous}, \projectname \ outperforms previous symbolic reasoning systems, including LPA \citep{chen-etal-2020-tabfact},  SASP \citep{ou-liu-2022-learning}, Binder \citep{cheng-etal-2023-binding}, and a state-of-the-art natural logic system \citep{aly-etal-2023-qanatver}, with a lead of \feverousaccleadfullSymbol \ accuracy points over the best performing baseline, Binder. Moreover, \projectname \ outperforms the highest-scoring neural entailment model, a classifier-version of the same language model used by \projectname, by \feverousaccleadfull \ accuracy points. \projectname outperforms further classification baselines in a few-shot setting, such as TAPAS \citep{herzig-etal-2020-tapas}, TAPEX \citep{liu-etal-2022-tapex}, and PASTA \citep{gu-etal-2022-pasta}.
We confirm the tabular reasoning capabilities of \projectname \ in a domain transfer setting to Tabfact \citep{chen-etal-2020-tabfact} without further training annotations, where our system performs competitively, improving on the strongest baseline by \tabfactaccleadfull \ accuracy points.
Our analysis reveals that \projectname's \ reading of numerals is more sensitive to numerical inaccuracies and the context of a claim (like quantifiers and adverbial modifiers) than a same-sized LLM baseline, reflecting the annotator guidelines of FEVEROUS more accurately. A qualitative discussion highlights attractive properties of \projectname \ and natural logic over existing SQL systems for veracity prediction over tabular evidence, such as its ability to model the NEI label directly.\footnote{Code at \url{https://github.com/Raldir/TabVer}} %

\section{Related Work}
\label{sec:related-work}

Symbolic reasoning systems for fact verification convert text into a logical form or executable program (SQL/LISP-style). They typically involve a neural component, either to rank viable candidate programs consisting of hand-crafted functions \citep{chen-etal-2020-tabfact} or via neural-symbolic models that generate programs directly \citep{liang-etal-2017-neural, ou-liu-2022-learning}. These programs are faithful explanations since the program's execution is the verdict. With the improved capabilities of large language models to generate code \citep{chen-etal-2021-evaluating}, \citet{cheng-etal-2023-binding, glenn-etal-2024-blendsql} explore the use of SQL, Python, and FOL to faithfully fact-check tabular claims, however, they only use proprietary models consisting of hundreds of billions of parameters. We show that \projectname \ outperforms these approaches (when controlled for the language model), which we attribute to the suitability
of natural logic to natural language in contrast to query languages like SQL.

The aforementioned symbolic executioners stand in contrast to the more prominent approach of using programs as features to neural systems, typically complemented by the original claim and table. For instance, LISP-style programs are used as a latent signal for a graph neural network \citep{shi-etal-2020-learn, zhong-etal-2020-logicalfactchecker, yang-etal-2020-program, gong-etal-2023-double}, and SQL queries and their executions are used as features to an LLM serving as a verdict classifier \citep{kong-etal-2024-opentab,zhang-etal-2024-reactable, wu-feng-2024-protrix}. \citet{wang-etal-2024-chainoftable} incrementally update an evidence table with LISP-style operations. Alternatively to symbolic integration into neural systems, \citet{chen-2023-large} produce natural language explanations using chain-of-thought prompting \citep{wei-etal-2022-chainofthought}. \citet{chen-2023-large} show that a 175B parameter GPT-3 model competes with fully supervised systems on tabular claims, yet its 6.7B variant performed only slightly above chance. This observation has been further confirmed by \citet{zhang-etal-2024-are} with Llama2-Chat-7B. Finally, large-scale instruction-tuning on tabular tasks has been explored \citep{zhuang-etal-2024-structlm, zhang-etal-2024-tablellama, liu-etal-2023-zero}, however they do not produce explanations. Conclusively, previous systems either rely on large proprietary models to achieve competitive performance %
or they sacrifice prediction explainability.

In contrast to these explicit meaning representations, \citet{angeli-manning-2014-naturalli} %
propose to extend the NatLog system of natural logic \citep{maccartney-manning-2007-natural, maccartney-manning-2009-extended} for textual inference, operating directly on natural language by \emph{comparing} texts in a premise with an associated hypothesis using set-theoretic relations. Thus, as a framework of flexible compositional inference, it circumvents the requirement to convert statements into rigid logical forms, and typically independently from one another. These favourable properties of natural logic inference have subsequently recently been explored for fact verification,
resulting in accurate predictions while maintaining transparency with plausible explanations \citep{krishna-etal-2022-proofver, aly-etal-2023-qanatver, strong-etal-2024-zeroshot}. \citet{aly-etal-2023-qanatver} and \citet{strong-etal-2024-zeroshot} exploit natural logic's operations on natural language by casting the operators into a question-answering framework to leverage recent advances of instruction-tuned language models.
This paper is the first attempt to extend %
natural logic inference for fact verification to the tabular domain. %

Finally, tabular question answering  \citep{jin-etal-2022-survey} is a common component to decompose a claim and reasoning processes.  
\citet{yang-zhu-2021-exploring} supplement the evidence with answers to questions generated via decomposition templates while \citet{suadaa-etal-2021-tabletotext} supplement the evidence with information from a table-to-text model. More recently, \citet{ye-etal-2023-large} use LLMs to decompose tables and questions. However, all three methods feed these modified tables into a pre-trained neural model \citep{herzig-etal-2020-tapas}, ultimately producing veracity predictions without explanations. Finally, even for textual evidence, most previous work that generates questions conditioned on the claim does not construct proofs from the answers \citep{rani-etal-2023-factify5wqa, fan-etal-2020-generating, jobanputra-2019-question}. %

\section{Method}

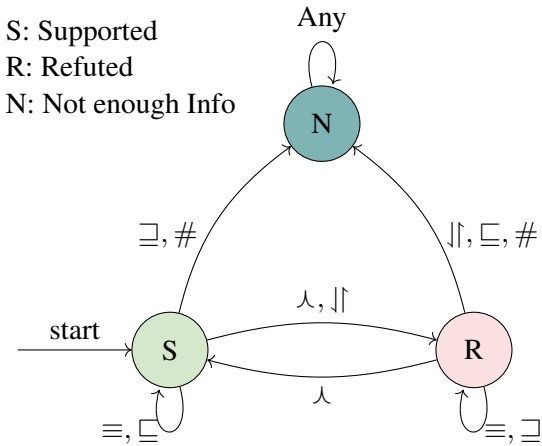
\begin{figure}[ht!]
\begin{tikzpicture}[
    node distance=6cm,
    state/.style={circle, draw, minimum size=1cm},
    ]
    
    \node[state,fill=celadon!50] (S) at (-2,-2) {S};
    \node[state,fill=babypink!50] (R) at (2,-2) {R};
    \node[state,fill=paperblue!50] (N) at (0,1) {N};
    
    \draw[->] (S) edge[loop below] node[left] {$\mynatop{equiv}, \mynatop{forwardentailment}$} (S);
    \draw[->] (R) edge[loop below] node[right] {$\mynatop{equiv}, \mynatop{reventailment}$} (R);
    \draw[->] (N) edge[loop above] node[above] {Any} (N);
    \draw[->] (S) edge[bend left=20] node[left,pos=0.4] {$\mynatop{reventailment}, \mynatop{independence}$} (N);
    \draw[->] (R) edge[bend right=20] node[right,pos=0.4] {$\mynatop{alternation}, \mynatop{forwardentailment}, \mynatop{independence}$} (N);
    \draw[->] (S) edge[bend left=15] node[above] {$\mynatop{negation}, \mynatop{alternation}$} (R);
    \draw[->] (R) edge[bend left=15] node[below] {$\mynatop{negation}$} (S);
    
    \draw[->] (-4,-2) -- node[above] {start} (S);
    
    \node[anchor=north east, align=left] at (-1, 2.5) {
        S: Supported\\
        R: Refuted \\
        N: Not enough Info
    };
\end{tikzpicture}
\caption{The finite state automaton (DFA), following natural logic inference \citep{angeli-manning-2014-naturalli}. The transitions in the DFA denote NatOps and the states the veracity labels. The final state on which the the proof terminates determines the overall veracity.}
\label{fig:natlog-dfa}
\end{figure}

Given a claim $c$ and a set of evidence tables $E$, the task is to predict a veracity label $\hat{y} \in $ \{\text{Supports}, \text{Refutes}, Not Enough Information (NEI)\}, and to accompany the prediction with an explanation. Since 
evidence might require arithmetic reasoning beyond the expressiveness of natural logic, as shown in Figure~\ref{fig:method-overview} with \emph{three municipalities}, \projectname's explanation is a proof $P = m_1, \ldots, m_l$, consisting of quintuples $m_i=(c_i, e_i, q_i, a_i, o_i)$, where $o_i$ describes the set-theoretic relation (NatOp) between a claim span $c_i$ and the result $a_i$ of arithmetic computations executed over relevant evidence $e_i$.  \projectname \ performs arithmetic reasoning steps in a question-answering framework, producing an arithmetic expression (ArithExp), with $a_i$ being the answer to a question $q_i$ for a claim span $c_i$ answered over evidence $e_i$. The sequence of operators $O=o_1, \ldots, o_l$ is then the input to a finite state automaton that specifies the claim's veracity label $\hat{y} = \text{DFA}(O)$. We follow the DFA for textual entailment described in \citet{angeli-manning-2014-naturalli}, shown in Figure \ref{fig:natlog-dfa}.

\begin{figure*}
\centering
\begin{tikzpicture}[
    box/.style={draw, minimum width=3cm, minimum height=2.5cm},
    smallcircle/.style={circle, minimum size=1cm, draw=black, line width=0.5pt},
    bigcircle/.style={circle, minimum size=2cm, draw=black, line width=0.5pt, distance=5pt},
    label/.style={font=\scriptsize, align=center},
    every node/.append style={font=\scriptsize},
    boxneg/.style={
    draw, 
    minimum width=3cm, 
    minimum height=2.5cm,
    path picture={
        \fill[bondiblue] (path picture bounding box.south west) rectangle 
            ([xshift=1.5cm]path picture bounding box.north west);
        \fill[lightgreen] ([xshift=1.5cm]path picture bounding box.south west) 
            rectangle (path picture bounding box.north east);
    },
}
]

\colorlet{lightgreen}{amethyst!30}

\tikzset{
  custom north west lines/.style={
    pattern={
      Lines[angle=45,distance=#1,line width=2pt]
    }
  },
  custom north west lines/.default=5pt
}

\node[box] (eq) {};
\node[bigcircle, fill=bondiblue] at (eq) {};
\node[bigcircle, custom north west lines=11pt, line width=0.2pt, pattern color=amethyst!70] at (eq) {};
\node[above=0mm of eq.north] {Equivalence ($\mynatop{equiv}$)};
\node[text width=3cm, below=1mm of eq.south, label] {\textcolor{bondiblue}{3 goals} $\mynatop{equiv}$ \textcolor{amethyst}{three goals}};

\node[box, right=2mm of eq] (fwd) {};
\node[bigcircle, fill=lightgreen] at (fwd) {};
\node[smallcircle, fill=bondiblue] at ([shift={(-2mm,-2mm)}]fwd.center) {};
\node[above=0mm of fwd.north] {Forward Entailment ($\mynatop{forwardentailment}$)};
\node[text width=3cm, below=1mm of fwd.south, label] {\textcolor{bondiblue}{99 goals} $\mynatop{forwardentailment}$ \textcolor{amethyst}{scored about a hundred goals}};

\node[box, right=2mm of fwd] (rev) {};
\node[bigcircle, fill=bondiblue] at (rev) {};
\node[smallcircle, fill=lightgreen] at ([shift={(-2mm,-2mm)}]rev.center)  {};
\node[above=0mm of rev.north] {Reverse Entailment ($\mynatop{reventailment}$)};
\node[text width=3cm, below=1mm of rev.south, label] {\textcolor{bondiblue}{2+ goals} $\mynatop{reventailment}$ \textcolor{amethyst}{3 goals}};

\node[box, right=2mm of rev] (alt) {};
\node[smallcircle, fill=bondiblue] at ([shift={(5mm,5mm)}]alt.center) {};
\node[smallcircle, fill=lightgreen] at ([shift={(-5mm,-5mm)}]alt.center) {};
\node[above=0mm of alt.north] {Alternation ($\mynatop{alternation}$)};
\node[text width=3cm, below=1mm of alt.south, label] {\textcolor{bondiblue}{5 goals} $\mynatop{alternation}$ \textcolor{amethyst}{has scored two goals}};

\node[boxneg, right=2mm of alt] (neg) {};
\node[above=0mm of neg.north] {Negation ($\mynatop{negation}$)};
\node[text width=3cm, below=1mm of neg.south, label] {\textcolor{bondiblue}{3 goals} $\mynatop{negation}$ \textcolor{amethyst}{never scored three goals}};

\foreach \x in {eq,fwd,rev,alt,neg}
    \node[anchor=north east] at (\x.north east) {$\mathbb{R}$};
\end{tikzpicture}
\vspace{-1.5em}
\caption{A set-theoretic view of the relationship between claim spans \textcolor{amethyst}{$c_i$} and numerical expressions in the evidence \textcolor{bondiblue}{$e_i$} when following an upper-bounded interpretation of numerals.} %
\label{fig:set-theoretic-definition}
\end{figure*}
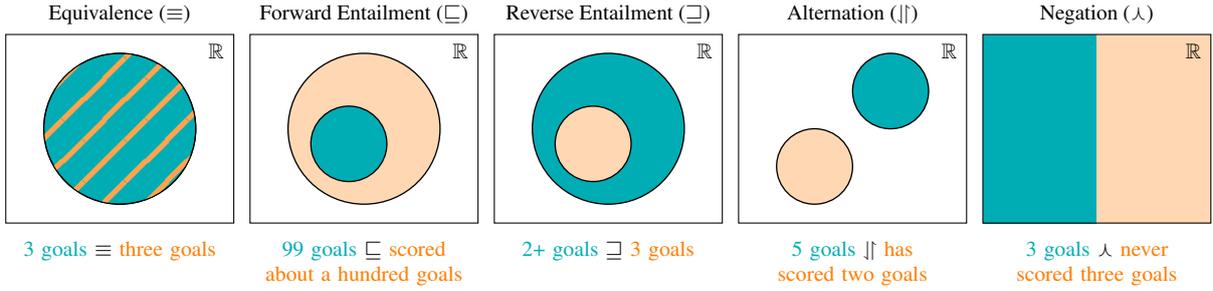

To enable the assignment of NatOps $o$ to ArithExps, we need to expand the set-theoretic definition of these operators. To this end, we first discuss the set-theoretic relationship for numerals that occur in claim and evidence %
without the need for further computation 
(Sec.~\ref{sec:set-theoretic-definition}). We subsequently expand this definition to %
ArithExps where arithmetic functions are applied to evidence, by mapping function executions on relevant evidence to numerical representations (Sec.~\ref{sec:arithmetic-functions-arithexp}).
\projectname \ produces its quintuples $(c_i, e_i, q_i, a_i, o_i)$ by first generating a question $q_i$ about a claim span $c_i$ that contains salient information (Sec.~\ref{sec:question-generation}). This question is answered using the evidence $E$ by producing a rationale, consisting of extracted evidence $e_i$, the execution of appropriate arithmetic functions on $e_i$, and the final answer $a_i$ (Sec.~\ref{sec:tabular-qa-natop}). Finally, a proof generation model $M_{P}$, trained on proofs containing ArithExps and associated NatOps following our set-theoretic definitions, assigns a NatOp $o_i$ to the claim-answer pair. %
\projectname \ follows  QA-NatVer \citep{aly-etal-2023-qanatver} for the proof generation process by selecting over multiple proof candidates.

\subsection{A Set-theoretic Perspective on Numerals}
\label{sec:set-theoretic-definition}
We first define a set-theoretic interpretation of the relationship between numerals in claim spans and evidence (or answers calculated on the evidence with ArithExps), within the context of natural logic. Specifically, we consider five set-theoretic relationships (NatOps) $o \in \{\equiv, \sqsubseteq, \sqsupseteq, \curlywedge, \downharpoonleft \! \upharpoonright, \}$\footnote{We do not define a mapping to the independence NatOp (\#) since it is applied when none of the other operators are predicted. Similarly to \citet{krishna-etal-2022-proofver} for textual relations, we observe that the cover NatOp occurs only very rarely, thus replacing it with the independence NatOp (\#).}. Figure~\ref{fig:set-theoretic-definition} shows examples of numerical expressions as evidence $e_i$ with the associated claim span $c_i$ for each NatOp. For instance, a claim span \emph{about a hundred goals} would generally be supported by the evidence \emph{99 goals} since the explicit adverbial modifier \emph{about} widens the scope of the numeral \emph{a hundred} to a larger set, including for instance \emph{99} and \emph{101}. However, even bare numerals can carry implicit meaning beyond the utterance itself, referred to as scalar implicatures \citep[][\textit{inter alia}]{grice-1975-logic}, and are subject to both semantics and pragmatics. This was also pointed out by \citet{maccartney-2009-natural} in the context of natural logic, but was ultimately ignored in their NatLog system.

Linguistic approaches to numerals typically consider an upper-bounded (exact) and a lower-bounded (at least) reading, which depend on several factors, such as whether the context is upward- or downward-monotone \citep{panizza-etal-2009-role}.\footnote{Downward-monotone environments are for instance negative environments, antecedent clauses of conditional constructions, restrictors of universal quantifiers \citep{spector-2013-bare}. Example for upward (downward) entailment: Messi (has not) scored 50 goals in a season.} Suitably, the effect of these environments on the entailment relationship between claim and evidence is modelled explicitly in natural logic via projection functions \citep{maccartney-2009-natural}, thus incorporating the different readings of numerals into a model of natural logic. Numbers typically follow an upper-bounded reading in an upward-monotone environment. An upper-bounded reading assumes that spans like \emph{2 goals} and \emph{5 goals} are mutually exclusive without covering the entire universe, which is, in our case, all natural numbers (cf.\ Appendix \ref{app:method-details}). Consequently, the terminology of natural logic can be extended in such environments, such that evidence spans like \emph{5 goals} aligned to claim spans with a strictly smaller number, like \emph{2 goals}, are assigned the alternation NatOp ($ \downharpoonleft \! \upharpoonright$). This reading is illustrated in Figure~\ref{fig:set-theoretic-definition}.
In contrast, downward-monotone environments typically cause a lower-bounded reading of numerals. Continuing with the above example, we see that in a downward-monotone environment the statement \emph{Everybody who scored 2 goals received a bonus} entails via forward entailment ($\mynatop{forwardentailment}$) the statement \emph{Everybody who scored 5 goals received a bonus}, since the numbers have an \emph{at least} interpretation without further specification, following \citet{panizza-etal-2009-role}. The resulting projection functions of upward- and downward-monotone environments on a numeral's reading are summarised in Appendix Table \ref{tab:tabver-projection-function}. %

Another component of a numeral's reading to consider is its \emph{pragmatic halo} \citep{lasersohn-1999-pragmatic}, where a number can represent a range of values due to the intended degree of approximation to its exact value in a specific context. A halo can be indicated explicitly with modifiers, as seen earlier (cf.\ \emph{about}), or defined implicitly. For instance, a claim like ``\emph{Messi scored a hundred goals in the 2010 season.}'' might be considered supported by evidence which states that he scored $101$ goals, in the context of an environment with low requirements of numerical precision, e.g.\ social media. In this example, the communicated content, \{100\}, is weaker than the asserted content, $101 \in \{100\} \cup H_{100}$, with $H_{100}$ being the pragmatic halo of $100$ as a set of (integer) numbers.\footnote{Note that this phenomenon is distinct from truth-conditional vagueness. %
While a sentence like ``\emph{Messi scored about 100 goals, he scored 102.}'' is semantically felicitous, ``\emph{Messi scored 100 goals, he scored 102}'' is not without explicitly correcting the first statement with a modifier like \emph{actually}, e.g.\ ``\emph{Messi scored 100 goals, actually he scored 102}" \citep{lauer-2012-pragmatics}.} However, the same evidence would lead to the claim's refutation in an environment where exactness is required, i.e.\ when $H_{100} = \emptyset$, e.g.\ statements in scientific articles.\footnote{\citet{lasersohn-1999-pragmatic} argues that the term \emph{exact} also leaves room for a non-empty pragmatic halo at times, e.g.\ in a statement such as \emph{Mary arrives exactly at 5 o'clock}, where deviations by milliseconds are permissible in most situations. We ignore this notion for simplicity.}
The size of the pragmatic halo typically increases with larger numbers, thus it becomes less necessary pragmatically to be precise \citep{woodin-etal-2024-largescale}. Therefore, \citet{vlachos-riedel-2015-identification} consider a fixed threshold on the absolute percentage error between numbers in a claim and evidence. Yet, in reality, the halo of a number is more dynamic: in decimal number systems, such as English, multiples of ten and five generally have a larger pragmatic halo than others due to the communicative tool of rounding \citep{woodin-etal-2024-largescale}. For instance, the claim that \emph{Messi scored 100 goals} while evidence states he scored 101 is more likely to be accepted than the reverse, since $101$ is not expected to be a rounded number, hence $|H_{100}|$ > $|H_{101}|$. %
Conveniently, the pragmatic halo can be expressed by natural logic via a projection to the entailment NatOps %
(e.g.\ Forward Entailment ($\mynatop{forwardentailment}$) in Figure~\ref{fig:set-theoretic-definition}) and is learned on annotated proof data (cf. Section \ref{sec:implementation-details}).

\subsection{Arithmetic Expressions}
\label{sec:arithmetic-functions-arithexp}
Since evidence $e_i$ is often stated in terms different than those needed to verify a textual claim, e.g.\ as seen in Figure~\ref{fig:method-overview}, where the number of municipalities, \emph{three}, is not found verbatim in the evidence, we introduce ArithExps, which map tabular evidence to numerals by executing arithmetic functions. 
ArithExps are function executions that produce an answer $a_i$ for a question $q_i$ to an associated claim span $c_i$ over relevant evidence $e_i$ from the table $E$. %
For the computation of $a_i$ we consider functions $\mathbb{F}: \mathbb{R}^n \rightarrow \mathbb{R}$ that take as input evidence $e_i$ and output a single numeral. The answer of the ArithExp is represented as the result of the computation prepended by the function's name: $\text{Name}(\mathbb{F}) \oplus \mathbb{F}(e_i)$, with $e_i \subseteq E$.\footnote{Despite the ArithExp's treatment as a numeral, the function name as part of $a_i$ is important since the semantics of a numeral varies between arithmetic functions (e.g.\ COUNT 5 versus COMP 5) and thus affect the comparison against claim span $c_i$.%
} Figure~\ref{fig:method-overview} shows the ArithExp \emph{SUM 12,238} (for the sum of the cells $3,945$, $1,126$, $1,363$, and $5,804$) as answer $a_i$ aligned to the claim span $c_i$ \emph{larger than 12,000}. To extend ArithExps to cover more complicated computations, we enable function composition, i.e.\ a function $\mathbb{G}$ as an input argument to $\mathbb{F}$. The ArithExp for function composition is the final computation, i.e.\ $\mathbb{F}$ for $\mathbb{F}(\mathbb{G}(E_i), \mathbb{G}(E_j),\cdots)$. 

\begin{figure}[ht!]
   \begin{tikzpicture}[
    box/.style={draw, rectangle, rounded corners, minimum width=2cm, minimum height=1cm},
    arrow/.style={->, >=stealth, thick},
    every node/.append style={font=\scriptsize},
]

    \definecolor{customblue}{RGB}{0,128,128}
    \definecolor{customred}{RGB}{255,0,0}
    
    \newcommand{\blue}[1]{\textcolor{customblue}{#1}}
    \newcommand{\red}[1]{\textcolor{customred}{#1}}

    \node[text width=8cm] at (0,0) (question) {\textbf{Question}: What is the total population of Ortegal in 2018?};

    \node[text width=5cm, minimum height=2cm] at (-1.4,-2) (table) 
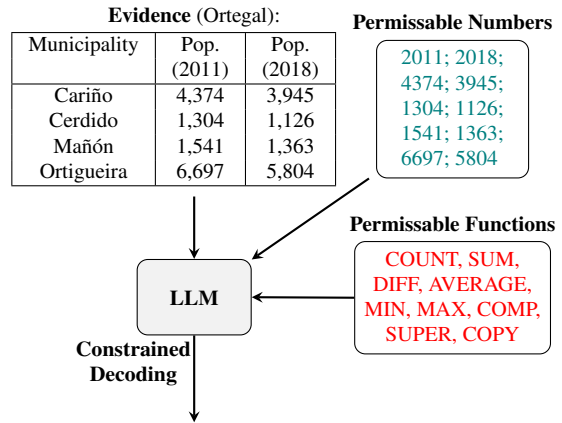
{
        {  
         \begin{tabular}{|c|c|c|}
            \hline
            Municipality & Pop. & Pop. \\
            &  (2011)  & (2018) \\
            \hline
            Cariño & 4,374 & 3,945 \\
            Cerdido &  1,304 & 1,126 \\
            Mañón & 1,541 & 1,363 \\
            Ortigueira & 6,697 & 5,804 \\
            \hline
        \end{tabular}
        }
    };
    \node[above=-0.2cm of table] {\textbf{Evidence} (Ortegal): };
    
    \node[box, align=left] at (2,-2) (numbers) {
        \scriptsize
        \blue{2011; 2018;} \\
        \blue{4374; 3945;} \\
        \blue{1304; 1126;} \\
        \blue{1541; 1363;} \\
        \blue{6697; 5804}
    };
    \node[above=0.0cm of numbers] {\textbf{Permissable Numbers}};
    
    \node[box, align=center] at (2,-4.5) (functions) {
        \scriptsize
        \red{COUNT, SUM,} \\
        \red{DIFF, AVERAGE,} \\
        \red{MIN, MAX, COMP,} \\
        \red{SUPER, COPY}
    };
    \node[above=0.0cm of functions] {\textbf{Permissable Functions}};
    
    \node[box, fill=gray!10, minimum width=1.5cm, minimum height=1cm, anchor=south, shift={(0,-3)}] at (table) (llm) {\textbf{LLM}};
    
    \node[align=center, text width=2cm, anchor=south, shift={(-0.8cm, -1.3cm)}] at (llm) (decoding) {\textbf{Constrained Decoding}};

    \node[align=left, text width=8cm,] at (0,-7) (extraction) {
        \textbf{Extraction}: In \blue{2018} Cariño had a population of \blue{3,945}, \\
        Cerdido \blue{1,126}, Mañón \blue{1,363}, and Ortigueira \blue{5,804}. \\[1ex]
        \textbf{Compute}: \red{Adding} \blue{3,945} \red{+} \blue{1,126} \red{+} \blue{1,363} \red{+}  \blue{5,804} = 12,238. \\
        \textbf{Answer ($a_i$)}: \underline{\red{SUM} \blue{12,238}} %
    };

    \draw[arrow] (table.south) -- (llm.north);
    \draw[arrow] ([xshift=-1.1cm, yshift=0]numbers.south) -- ([xshift=0cm, yshift=0.5cm]llm.east);
    \draw[arrow] (functions.west) -- (llm.east);
    \draw[arrow] (llm.south) --  ([xshift=-1.4cm, yshift=0]extraction.north);
    \end{tikzpicture}
    \vspace{-1.5em}
    
    \caption{Tabular question answering via a rationale that produces the answer $a_i$ via an ArithExp. An LLM jointly extracts relevant information from table cells and executes appropriate functions. %
    The generation is constrained to permissible functions and to numbers that appear in the evidence to alleviate hallucinations.}
    \label{fig:arith-exp-generation}
\end{figure}

The full list of permissible functions $\mathbb{F}$ we consider is shown in Figure~\ref{fig:arith-exp-generation}. In addition to the functions \emph{count, sum, diff, average, min}, and \emph{max}, we consider comparative functions as a separate function class. Comparatives could be modelled by the \emph{diff} function, thus subtracting quantities between relevant arguments. However, we represent them as a unique ArithExp since they serve a different semantic function in relation to a claim span $c_i$. The comparative ArithExp can be used for both imprecise (e.g.\ \emph{Person X had more votes than Person Y})  as well as precise comparisons (e.g.\ \emph{Person X had 5000 votes more than Person Y}), since the difference in quantity is indicative of both polarity and magnitude. Finally, to cover the base case where all relevant information is already contained in $e_i$ (i.e.\ no computation is required, cf.\ Sec. \ref{sec:set-theoretic-definition}), we include a copy function. %

\subsection{Question Generation}
\label{sec:question-generation}
 We generate questions that can be linked directly to salient parts of a claim $c_i$, as seen in Figure \ref{fig:method-overview}. For instance, the question \emph{What is the total population of Ortegal in 2018} directly corresponds to the claim span \emph{larger than 12,000}.  %
We use a fine-tuned large language model $M_{QG}(c, T)$, which takes a claim $c$ and a prompt template $T$ as input and autoregressively generates a collection of questions  $q_1 \ldots q_l$, along with their corresponding targeted claim spans. %
The output is formatted as a list of questions and claim spans 1. [q1] [c1] 2. [q2] [c2]$\cdots$. To ensure that the generated claim span occurs verbatim in the claim, we employ constrained decoding to restrict the sampling of $c_i$ to spans of the claim $c$ (including $c$ itself). Thereby we prevent the model from introducing words or phrases that are not present in the claim, a behavior we observed even after fine-tuning. Additionally, we use constrained decoding to enforce the enumeration format defined in the prompt above for generating multiple questions jointly. By conditioning the generation of questions on previously generated ones, we can improve coverage of salient information in the claim and reduce the likelihood of redundant or repetitive questions \citep{fan-etal-2020-generating}.

\subsection{Tabular QA with ArithExps}
\label{sec:tabular-qa-natop}
The next step is to answer a generated question $q_i$ %
using information from the evidence tables $E$ whilst using only permissible functions $\mathbb{F}$ to obtain the answer $a_i$ and the ArithExp.
We use a fine-tuned language model M$_{QA}$ which takes as input a question $q_i$, associated with claim span $c_i$, and evidence tables $E$, and generates a rationale $J$ consisting of three parts: table-to-text conversion to extract relevant evidence $e_i$ from table cells in $E$; the execution of relevant arithmetic functions on $e_i$; and the answer representation $a_i$. The components of the rationale $J$ are generated jointly in an autoregressively fashion:
\begin{align}
M_{QA}(J \mid q_i, E) &= \underbrace{\prod_{u=1}^{U}p_{\theta}(s_u |s_{<u}, q, E)\nonumber}_{p_{\theta}(e_i|  q_i, E)}\\
&\cdot \underbrace{\prod_{m=1}^{M}p_{\theta}(f_m |f_{<m}, q_i, e_i, E)\nonumber}_{p_{\theta}(\mathbb{F}| q_i, E, e_i)}
\end{align}
with $s_u$ and $f_m$ being decoded tokens over the extracted evidence $e_i$ from cells in $E$ and arithmetic functions $\mathbb{F}$, %
respectively, and $\theta$ being the parameterization of $M_{QA}$. The components of $J$ are generated using a different decoding scheme, illustrated in Figure~\ref{fig:arith-exp-generation}. To avoid hallucinations in the extracted evidence $e_i$, the probability $p_{\theta}(e_i|q_i, E)$ is set to $0$ for sequences where numbers in the generated sequence do not occur in any table cells of $E$. The generation of $p_{\theta}(\mathbb{F}|q_i, E, e_i)$ is constrained such that the first generated word needs to be a trigger word for one of the permissible functions $\mathbb{F}$ (such as \emph{Adding} for \emph{sum}), followed by the function itself. Finally, the answer $a_i$ is constructed deterministically from the rationale to align with the representation of ArithExps as described in Section \ref{sec:arithmetic-functions-arithexp}. %
As shown in the example Figure~\ref{fig:arith-exp-generation}, the extraction of the population numbers is constrained to numbers occurring in the evidence table, such as \emph{2018; 3,945; 1,126; 1,363; and 5,804} (blue), the arithmetic computation is constrained to start with the trigger word \emph{Adding} (red), followed by the execution of the associated function, with the final answer $a_i$ being \emph{SUM 12,238}.

If no function is considered relevant to further process the evidence $e_i$, the model $M_{QA}$ outputs \emph{N/A} after the extraction of evidence and subsequently does not return an ArithExp. %
If the evidence tables do not contain any relevant information to answer $q$, then the model returns \emph{N/A} as the relevant evidence $e_i$ which is mapped to an independence NatOp (\#), leading to an NEI verdict prediction according to the DFA (cf.\ Figure \ref{fig:natlog-dfa}). Parts of a claim that do not require separate questioning (such as \emph{In 2018} in Figure~\ref{fig:method-overview}) are assumed to be contained in extracted evidence for answering questions about claim $c$. QA-NatVer's span alignment algorithm aligns these claim spans to extracted evidence $e_i$ from all the questions $q_1 \ldots q_l$ to the claim $c$.

\section{Evaluation}

\subsection{Data}
\label{sec:data}
 
\paragraph{FEVEROUS.} We train and evaluate models on the tabular subset of FEVEROUS \citep{aly-etal-2021-feverous}, i.e., the claims where all evidence elements across all evidence sets are cells from tables. FEVEROUS consists of complex claims and tables with irregular structures. %
To focus on the natural logic-based tabular fact verification component of fact-checking, we use gold evidence tables (i.e.\ not ones selected via a retrieval system from a knowledge source) throughout our experiments. %
The resulting dataset consists of $2,011$ claims, with 35\%, 61.7\%, and 3.2\% being supported, refuted and NEI claims, respectively (cf.~Appendix \ref{tab:feverous-tables-stats}). Out of the $2,011$ claims, $521$ are labelled as requiring numerical reasoning. 

Models are trained on $64$ FEVEROUS instances, selected uniformly from its training data. The veracity labels in the resulting training data are thus similarly imbalanced as the FEVEROUS development data. Additionally, to train \projectname, we manually annotated these training instances with rationales and natural logic proofs. These proofs contain ArithExps as defined in Section \ref{sec:set-theoretic-definition}. The training distribution of arithmetic functions is also imbalanced. For details, see Appendix \ref{app:dataset-details}.

\paragraph{TabFact.} We further evaluate models trained on FEVEROUS in a domain transfer scenario on TabFact \citep{chen-etal-2020-tabfact}, without further training on the latter. Contrary to FEVEROUS, TabFact only contains two veracity labels: Supported and Not Supported, the latter covering both refutations and NEI instances. TabFact contains only well-structured tables; the first row is always the table header. TabFact is designed to be evaluated on gold evidence tables $E$. We evaluate methods on its development set, consisting of $12,851$ claims with evenly distributed labels, out of which $4424$ are simple (R1) and $8427$ complex claims (R2). %

\subsection{Baselines}

We compare \projectname \ against strong baselines that can be categorized into two classes: (i) classifiers that predict a veracity label without symbolic mechanisms or explanation production; and (ii) symbolic reasoning models that produce faithful explanations.

\paragraph{Classification models.} \textbf{DeBERTa+NLI} is a DeBERTaV3 model \citep{he-etal-2023-debertav3} fine-tuned on multiple NLI tasks. %
\textbf{PASTA} \citep{gu-etal-2022-pasta} is a DeBERTaV3 model further pre-trained on different tabular operations. \textbf{TAPAS} \citep{herzig-etal-2020-tapas} is a transformer pre-trained on tabular data with additional table-aware positional embeddings. \textbf{TAPEX} \citep{liu-etal-2022-tapex} is based on BART \citep{lewis-etal-2020-bart}, pre-trained as an SQL executor and fine-tuned on tabular data via table linearization. We follow typical encoder-only fine-tuning, where a linear transformation from embeddings to veracity labels is jointly optimized with the pre-trained model itself.  %
Furthermore, we evaluate several LLMs, including \textbf{Llama2-Chat-7B} \citep{touvron-etal-2023-llama} and \textbf{MistralOrca-7B} \citep{jiang-etal-2023-mistral}. We fine-tuned the LLMs via LoRA \citep{hu-etal-2022-lora}. %

\paragraph{Symbolic reasoning models.} We compare against \textbf{LPA} \citep{chen-etal-2020-tabfact}, a LISP-style program synthesis algorithm with hand-crafted functions and trigger words to prune the search space. It incorporates a fine-tuned transformer to rank candidate programs. \textbf{SASP} \citep{ou-liu-2022-learning} is built on top of Neural symbolic machines \citep{liang-etal-2018-memory} and considers both lexical and structure features to constrain program candidates and further uses TaBERT \citep{yin-etal-2020-tabert} and an LSTM \citep{hochreiter-schmidhuber-1997-long} for program generation. We also consider \textbf{Binder} \citep{cheng-etal-2023-binding}, an approach that uses LLMs to map tabular claims to SQL queries and to execute specific API calls embedded in the queries on tables. To maintain comparability with \projectname, Binder uses MistralOrca-7B as the LLM. If no viable program can be found for a given claim, LPA, SASP, and Binder fall back to an NEI/Not Supported prediction. \textbf{QA-NatVer} \citep{aly-etal-2023-qanatver} constructs natural logic inference proofs by casting natural logic operators into a question-answering framework. We linearize the evidence table and use the Flan-T5 3B backbone \citep{chung-etal-2024-scaling}.

\subsection{Implementation Details}
\label{sec:implementation-details}

\paragraph{Claim Decomposition.} A FEVEROUS claim typically contains multiple factual statements which might not all be covered by the annotated evidence tables, since its annotations are only required to be sufficient (but not necessarily complete) to reach a verdict. Subsequently, the annotation for a refuted claim might lack evidence for some other parts of the claim, resulting in erroneous NEI predictions. Thus, a FEVEROUS claim $c$ is decomposed into a list of sub-claims $C$, such that each sub-claim is an atomic statement contained in the claim. We use a language model $M_{D}$, fine-tuned on manually annotated decompositions of the same FEVEROUS training instances described above. During inference, the sub-claims are enumerated following the question generation decoding scheme. The decomposition prompt is shown in Appendix \ref{app:impl-details}.
The predictions for each subclaim are aggregated into a final verdict $\hat{y}$ via deterministic rules: %
\begin{align*}
\hat{y} &= \text{Supp} && \text{iff } \forall c \in C. \text{DFA}(O_c) = \text{Supp}; \\
\hat{y} &= \text{Ref} && \text{iff } \exists c \in C. \text{DFA}(O_c) = \text{Ref}; \\
\hat{y} &= \text{NEI} && \text{iff } \not \exists c \in C. \text{DFA}(O_c) = \text{Ref} \\
&&& \land \not \forall c \in C. \text{DFA}(O_c) = \text{Supp}. \nonumber
\end{align*}
Thus a claim $c$ is supported iff all subclaims are supported by evidence, and refuted iff there exists a sub-claim that is refuted. Not enough information is predicted in all other cases. Since these rules are deterministic, the final prediction remains faithful.
See Figure \ref{fig:proof-decompositon-example} for an example.

We use the same decomposition for \projectname \ as well as all symbolic reasoning baselines we consider to maintain comparability. Classification models use the original claim as input instead, since the impact of evidence incompleteness is expected to be minimal and decomposition can lead to error propagation. %
With the exception of \citet{wang-shu-2023-explainable}, who represent a claim as a conjunction over subclaims, the aggregations over verdicts of parts of a claim are executed via neural mechanisms and thus do not guarantee faithfulness \citep{chen-etal-2022-loren, zhao-etal-2024-pacar}.

\begin{table*}[ht!]
\resizebox{0.95\linewidth}{!}{
	\begin{tabular}{ l l|c|c|c|c|c} 
		\toprule
            & & \multicolumn{2}{c}{Full} & \multicolumn{2}{c}{Numerical} & Execution Found \\
            \cmidrule(lr){3-4} \cmidrule(lr){5-6} \cmidrule(lr){7-7} 
		  & & Accuracy & Macro F$_1$ & Accuracy &  Macro F$_1$ & (\%) \\
		\hline
            & Majority Baseline & 61.7  & 20.5 & 64.8 &  21.6 & -- \\
            \hline
            \textbf{Classific.}
           & DeBERTav3 & 53.9$_{0.7}$ & 36.8$_{0.4}$ & 55.6$_{0.6}$ & 36.0$_{1.3}$ & --\\
           & PASTA   & 54.6$_{2.8}$ & 34.1$_{0.4}$ & 55.3$_{4.3}$ & 32.6$_{1.4}$ & --  \\
	    & TAPAS    & 53.6$_{7.6}$ & 35.9$_{4.1}$ & 52.9$_{7.3}$ & 33.8$_{3.4}$ &  --  \\
	    & TAPEX   & 53.6$_{1.5}$ & 34.0$_{0.9}$ & 52.8$_{3.4}$ & 32.9$_{2.1}$ &  --  \\
            & Llama2-Chat-7B & 56.0$_{4.0}$ & 30.9$_{1.6}$ & 55.0$_{6.1}$ & 30.9$_{2.5}$ & -- \\
            & MistralOrca-7B & 68.0$_{1.1}$ & 45.4$_{4.4}$ & 64.5$_{3.2}$ & 43.6$_{3.0}$ & --\\
		\hline
            \textbf{Symbolic}
            & LPA (w/o decomp) & 31.6$_{0.4}$ & 27.5$_{0.5}$ & 37.3$_{0.7}$ & 28.1$_{0.9}$ & 54\% \\
            & LPA & 21.8$_{0.1}$ & 21.4$_{0.2}$ & 22.3$_{0.4}$ & 21.3$_{0.4}$ & 41\%\\
            & SASP (w/o decomp.) & 52.9$_{2.6}$ & 29.8$_{1.8}$ & 55.1$_{3.4}$ & 29.3$_{1.9}$ & 98\% \\
            & SASP & 58.8$_{0.8}$ & 29.6$_{0.8}$ & 61.5$_{1.2}$ & 29.4$_{0.8}$ & 95.2\% \\
            & Binder (w/o decomp.) & 60.9$_{1.2}$ & 38.0$_{1.3}$ & 61.0$_{1.6}$ &  40.1$_{2.2}$ & 95.7\% \\
            & Binder & 62.7$_{1.4}$ & 37.3$_{1.3}$ & 63.7$_{1.8}$ & 39.3$_{1.6}$ & 95.4 \%  \\
            & QA-NatVer &  54.0$_{1.1}$ & 34.8$_{0.2}$ & 52.6$_{1.6}$ & 28.9$_{0.3}$ & 100\% \\
		\hline
            \textbf{\projectname} & BART0  & 69.9$_{0.3}$ & 49.4$_{0.9}$ &  66.7$_{0.3}$ &  42.4$_{0.8}$  & 100\%  \\
            & FlanT5-xl & \textbf{71.4}$_{0.5}$ & \textbf{51.0}$_{0.5}$ & \textbf{70.1}$_{1.3}$ & \textbf{45.8}$_{0.3}$  & 100\% \\
		\bottomrule
	\end{tabular}}
	\caption{Verdict accuracy and macro-averaged F$_1$ on FEVEROUS. %
    \emph{Numerical} reports scores exclusively on the subset of claims involving numbers. %
    \emph{Execution found} is the proportion for which a program or proof was found.}
\label{tab:results-feverous}
\end{table*}

\paragraph{Experimental Setup.} %

We do not consider a validation set for hyperparameter-tuning, following the real-world few-shot learning setting of \citet{alex-etal-2021-raft} and instead use largely default hyperparameters. \projectname \ fine-tunes the question generation model $M_{QG}$, the question answering model $M_{QA}$, and the proof generation model $M_{P}$ on annotated handcrafted rationales and proofs described in Section \ref{sec:data}.  $M_{QG}$, $M_{QA}$, and and the claim decomposition model $M_{D}$ are MistralOrca-7B models, fine-tuned using LoRA \citep{hu-etal-2022-lora}. We use the proof generation model $M_{P}$ of \citet{aly-etal-2023-qanatver}. Specifcially, we fully fine-tune a FlanT5-3B parameter model and a smaller BART0 model (406M parameters) \citep{lin-etal-2022-unsupervised} as $M_{P}$ to measure the accuracy of \projectname  \ across model sizes. While it would be of interest to simplify \projectname  \ by using MistralOrca-7B (or another powerful LLM) for all components, the implementation in \citep{aly-etal-2023-qanatver} currently only supports the training of encoder-decoder models, following \citet{liu-etal-2022-fewshot}. Furthermore, while $M_{QG}$, $M_{QA}$, and $M_{D}$ require language generation, the proof generation model $M_P$ of \citet{aly-etal-2023-qanatver} solves a discriminative task (answering binary/ternary questions), for which encoder-decoders have shown to be competitive to decoder-only models on smaller scale (i.e. $\leq$ 11B parameters) \citep{chia-etal-2024-instructeval}.  Hyperparameters, prompts, and implementation details of \projectname and baselines are shown in Appendix~\ref{app:impl-details}. Results are averaged over five runs with the standard deviation indicated. In all other cases, results are reported using default seed $42$.

\section{Results}

 \begin{table*}[ht!]
\resizebox{1\linewidth}{!}{
	\begin{tabular}{ l l |c|c|c|c|c|c|} 
		\toprule
            
            & & \multicolumn{2}{|c|}{Full} &  \multicolumn{2}{|c|}{R1} & \multicolumn{2}{|c|}{R2} \\
            \cmidrule(lr){3-4} \cmidrule(lr){5-6} \cmidrule(lr){7-8} 
		& & Accuracy & Macro F$_1$ & Accuracy & Macro F$_1$ & Accuracy &  Macro F$_1$ \\ 
             \hline
            \textcolor{darkgray}{Full Supervision} & \textcolor{darkgray}{LPA} &  \textcolor{darkgray}{65.2} & \textcolor{darkgray}{64.2}  & \textcolor{darkgray}{77.6} & \textcolor{darkgray}{77.5} & \textcolor{darkgray}{57.4} & \textcolor{darkgray}{55.6} \\
            & \textcolor{darkgray}{SASP} & \textcolor{darkgray}{74.7} & \textcolor{darkgray}{74.7} & \textcolor{darkgray}{86.1} & \textcolor{darkgray}{86.1} & \textcolor{darkgray}{68.9} & \textcolor{darkgray}{68.9} \\
           &  \textcolor{darkgray}{TAPAS} & \textcolor{darkgray}{82.1} & \textcolor{darkgray}{82.0 } & \textcolor{darkgray}{92.8} & \textcolor{darkgray}{92.8} & \textcolor{darkgray}{76.5} & \textcolor{darkgray}{76.4} \\
             \hline
            \textbf{Classific.} & DeBERTav3 & 50.7$_{0.4}$ & 49.7$_{1.3}$ & 50.8$_{0.7}$ & 49.5$_{1.7}$ & 50.6$_{0.2}$ & 49.8$_{1.1}$ \\
            & PASTA  & 50.4$_{0.6}$ & 46.1$_{5.6}$ & 50.6$_{1.1}$ & 46.4$_{6.1}$ & 50.4$_{0.5}$ & 45.9$_{5.4}$ \\
		& TAPAS  & 53.9$_{5.9}$ & 53.0$_{6.8}$ & 58.8$_{10.6}$ & 58.5$_{11.3}$ & 51.3$_{3.6}$ & 49.9$_{4.7}$ \\
		& TAPEX  & 49.7$_{4.3}$ & 44.3$_{5.1}$ & 49.5$_{3.4}$ & 47.6$_{3.8}$ & 49.8$_{2.9}$ & 43.3$_{2.9}$ \\
            & Llama2-Chat-7B & 51.2$_{1.6}$ & 47.3$_{4.2}$ & 51.5$_{2.5}$ & 47.8$_{4.3}$ & 51.1$_{1.2}$ & 47.0$_{4.2}$ \\
            & MistralOrca-7B & 60.6$_{3.1}$ & 58.1$_{6.0}$ & 67.2$_{4.2}$ & 65.9$_{5.8}$ & 57.2$_{2.6}$ & 53.3$_{6.4}$ \\ 
		\hline
            \textbf{Symbolic} & LPA & 59.4$_{1.4}$ & 57.9$_{1.4}$ & 70.4$_{2.5}$ & 70.3$_{2.5}$ & 53.8$_{0.9}$ & 50.2$_{1.0}$ \\
            & SASP & 48.7$_{2.8}$ & 45.1$_{2.9}$ & 50.7$_{3.0}$ & 47.5$_{2.0}$ & 47.7$_{3.0}$ & 43.8$_{3.7}$ \\ 
            & Binder & 65.1$_{1.0}$ & \textbf{65.1$_{1.0}$}  & \textbf{76.9$_{0.6}$} & \textbf{76.9$_{0.6}$} & 59.1$_{1.3}$ & 59.1$_{1.3}$\\
            & QA-NatVer & 50.9$_{0.1}$ & 43.6$_{0.3}$ & 52.7$_{0.2}$ & 49.8$_{0.1}$ & 49.9$_{0.1}$ & 49.1$_{0.2}$ \\
            \hline
            \textbf{\projectname} & BART0  & 62.8$_{0.8}$ & 62.3$_{0.9}$ &  71.1$_{1.0}$ & 71.1$_{1.1}$ & 58.6$_{0.6}$ & 57.5$_{0.9}$ \\
            & Flan-T5-xl & \textbf{65.6}$_{0.3}$ & 64.8$_{0.6}$ & 72.6$_{0.5}$ & 72.2$_{0.6}$ & \textbf{62.1}$_{0.4}$ & \textbf{60.8}$_{0.9}$ \\
		\bottomrule
	\end{tabular}}
	\caption{Verdict accuracy and macro-averaged F$_1$ in a transfer scenario on TabFact, when trained on FEVEROUS. R1: Tabfact's subset of simple claims. R2: TabFact's subset of complex claims.}
 \label{tab:results-feverous-TabFact}
\end{table*}

\paragraph{FEVEROUS.} Results on FEVEROUS are shown in Table \ref{tab:results-feverous}, reporting both accuracy and macro average F$_1$ due to the dataset's label imbalance. \projectname \ outperforms all baselines both on the full dataset, as well as the numerical reasoning subset by \feverousaccleadfull \ and \feverousaccleadnumerical \ accuracy points, respectively. We see similar differences in terms of F$_1$, with a lead of \feverousfleadfull \ points. Except for the LLM MistralOrca-7B baseline, all classification models perform poorly in a few-shot scenario on FEVEROUS. Llama2-Chat-7B model's surprisingly poor performance confirms previous observations on few-shot tabular fact-verification \citep{chen-2023-large, zhang-etal-2024-are, zhang-etal-2024-tablellama}. In addition to the classification baselines being outperformed by \projectname, they lack transparency and faithful explanations.
To highlight \projectname's data efficiency, we compare it against a fully supervised TAPAS classification model trained on $18,836$ tabular FEVEROUS claims, where it achieves an accuracy score of $73.0$, performing only $1.6$ accuracy points better than \projectname. %

While symbolic reasoning baselines provide faithful explanations, their performance is substantially worse than \projectname. Symbolic reasoning systems that construct semantic representations are unable to handle diverse and complex tabular structures (e.g. nested table headers) as present in FEVEROUS. For instance, the %
rule-based LPA approach finds a suitable program only for 41\% of claims. The accuracy for claims where LPA finds a program is $55.8$ points, improving by $25.6$ points on its overall performance but still being outperformed substantially by \projectname. While the rate of executable programs is much higher for SASP and Binder due to the generation of programs being neural-guided, the overall performance is worse than \projectname , with a difference of \feverousaccleadfullSymbol \ accuracy points for the best performing symbolic baseline, Binder.
Finally, QA-NatVer has a 100\% execution rate due to its flexibility by operating on natural language similarly to \projectname, however, the difficulty of aligning linearized evidence to claims and the lack of arithmetic reasoning capabilities result in low scores. Interestingly, the symbolic baselines perform better or comparably on the numerical subset than on the full dataset, while we observe the opposite for the majority of classification models and natural logic-based approaches, confirming the difficulty for these meaning representations to model complex textual claims correctly. %

\paragraph{TabFact.}
Results in a domain-transfer scenario without TabFact training data are shown in Table~\ref{tab:results-feverous-TabFact}. \projectname \ still remains competitive with our baselines with an accuracy lead of  \tabfactaccleadfull \ accuracy points and an F$_1$ of \tabfactfleadfull \ points worse than the best baseline (Binder).
The performance against the symbolic reasoning systems is particularly noteworthy since LPA and SASP have been designed specifically for TabFact, and Binder's SQL parsing excels at well-structured tables. Subsequently, LPA, SASP, and Binder find viable programs more frequently than on FEVEROUS, with 78\%, 99.8\%, and 100\%, respectively. Binder performs the best out of all baselines, outperforming \projectname \ particularly on simple claims (R1) that do not require complex reasoning to predict correctly. Yet, on complex claims (R2) \projectname \ performs better than Binder. %
Binder's performance discrepancy between FEVEROUS and TabFact is noteworthy, %
highlighting a fundamental limitation to previous approaches when applied to diverse tables, which \projectname \ successfully addresses.

Training classification baselines, such as TAPAS, on Tabfact's $92,283$ training samples, using the same experimental setup, %
results in scores substantially outperforming all considered models (82.1 accuracy points). In contrast, TAPAS achieves a score barely above random in our transfer setting (53.9 accuracy points) since the small training size is insufficient for fine-tuning the model to the task and learning the linear transformation described in Section \ref{sec:implementation-details}. This problem is exemplified with TAPEX, as it is pre-trained only on SQL queries, necessitating substantial data during fine-tuning to learn a mapping to natural language. Compared to fully-supervised symbolic systems, \projectname \ remains competitive to LPA with an accuracy lead of $0.4$ points, but falls behind SASP substantially with an accuracy difference of $9.1$ accuracy points.

\paragraph{Reading of numerals.}
We further analyse \projectname's reading of numerals by isolating its ability to consider the context of numbers mentioned in a claim.\footnote{The ability of models to make pragmatic inferences has been explored in \citet{jeretic-etal-2020-are}, however, their dataset was constrained to a minimal scenario with four numbers (2, 3, 10, 100) and two quantifiers (\emph{some}, \emph{all}). Importantly, while their dataset focuses on correctness, our goal is instead to probe a model's reading of numerals.} We automatically construct a diverse probing dataset that considers variations of numbers in supported claims by adding numerical inaccuracies, rounding numbers, adding modifiers (i.e.,~\emph{approximately}, \emph{about}, \emph{around}), and adding cardinal determiners (i.e.,~\emph{at most}/\emph{least}). We measure the proportion between veracity predictions correctly labelled as supported and veracity predictions that remain supported after an inserted numeric variation. The probing dataset consists of $1638$ claims. For a detailed description of the constructed variations see Appendix \ref{app:reading-numeral-details}.

Table \ref{tab:pragmatic halo} shows the results of the probe for \projectname, Binder, and the MistralOrca-7B classification baseline. 
\projectname \ is substantially more sensitive to small numeric inaccuracies. Only for 36.3\% is the claim's prediction maintained when adding $1$ to the original number, compared to 63.4\% and 57.7\% for Binder and the classifier, respectively. This trend is also observed for relative numerical inaccuracies and rounded numbers. We argue \projectname's behaviour is more representative of its training data, since FEVEROUS instances are annotated to be refuted if numbers mentioned without modifier do not match exactly due to the guidelines given to annotators \citep{aly-etal-2021-feverous}. 
In contrast, when adding explicit modifiers, we observe that \projectname \ maintains its prediction more frequently than our baseline, with 57\% versus 40.6\% and 56.4\% for the classifier and Binder, respectively. Finally, \projectname's more nuanced reading of numerals is also seen for cardinals: while the classifier cannot differentiate between incorrect cardinal determiners (e.g.~ $12$ being modified to \emph{at most 10} and changing the veracity label, and \emph{at most 15} while preserving it), both \projectname \ and Binder differentiate between the two. Yet, Binder overall favours the prediction of supported, due to the answer-biased voting strategy deployed by \citet{cheng-etal-2023-binding}.%

\begin{table}[ht!]
\resizebox{0.9\linewidth}{!}{
	\begin{tabular}{ l|c c c} 
		\hline
		   & Class. & Binder & \projectname \\
     \hline
        Inaccuracy $\Delta$ +1 & 63.4\% & 57.7\% & 36.3\%  \\
	Inaccuracy $\Delta$ 2\% & 42.7\%  & 38.4\% & 29.1\% \\
	Inaccuracy $\Delta$ 10\% & 37.8\% & 38.4\%  & 26.3\% \\
        Inaccuracy $\Delta$ 25\%  & 31.7\% & 51.9\% & 23.6\% \\
    \hline
        Rounding  & 33.3\% & 47.4\% & 30.3\% \\
        Modifiers (e.g.~\emph{about}) & 40.6\% & 56.4\% &57.0\%\\ 
        Cardinal (incorrect) & 32.9\% & 53.8\% & 31.8\% \\ 
        Cardinal (correct) & 37.8\%& 78.8\% & 49.1\% \\ 
        \bottomrule
	\end{tabular}}
	\caption{Probing how modifications to numbers impact veracity predictions.%
 We report the proportion of claims for which a veracity prediction correctly labelled as supported does not flip after an edit to a claim's numeral. We consider absolute and relative numerical inaccuracies, approximations, explicit modifiers, and cardinals. }
\label{tab:pragmatic halo}
\end{table}

\paragraph{Correctness of ArithExps.}
To assess the quality of natural logic proofs with invoked ArithExps, we randomly select $160$ FEVEROUS samples and annotate the arithmetic functions required to reach the correct verdict. We compare these annotations with functions identified by \projectname as well as LPA's and Binder's programs. %
As seen in Table \ref{tab:discussion-correctness-arithexps}, the overall accuracy of \projectname's arithmetic function calls outperforms the LPA baseline with $76.0$ versus $43.8$ accuracy points, and is comparable with Binder's score of $76.5$. The largest performance lead for Binder is observed for the count function, whereas \projectname \ is more accurate at comparisons.

\begin{table}[ht!]
\centering
\resizebox{0.80\linewidth}{!}{
	\begin{tabular}{ l|c|c|c} 
		\hline
		   & LPA & Binder & \projectname \\
		\hline
            Overall & 43.8 & 76.5 & 76.0 \\ %
            \hline
		  Filter/Copy  & 41.7 & 85.6 &   90.4 \\ %
		  Comparisons & 33.3 & 0.0 &  25.0 \\
		Count & 75.0 & 85.7 &  46.4 \\
		Sum &  0.0 & 100.0 &  100.0 \\
		Diff  & 0.0 & 0.0 &  16.6 \\
            Min/Max & 0.0 & 0.0 & 0.0 \\
		\hline	
	\end{tabular}}
	\caption{Evaluation of arithmetic functions incorporated in \projectname's proofs, compared to LPA and Binder.}
	\label{tab:discussion-correctness-arithexps}
\end{table}

\paragraph{\projectname \ Ablation.}
Table \ref{tab:ablation} shows an ablation study of \projectname's components. We see a substantial performance decline when removing the generation of ArithExps, dropping accuracy on the numerical subset by $10.0$ accuracy points. When additionally removing the extracted evidence $e_i$ from the rationale and instead falling back to table linearization, 
we observe performance comparable to QA-NatVer, as expected. In line with our expectations, a major accuracy drop is observed on the full FEVEROUS data since the extraction and formatting of evidence is particularly useful for non-arithmetic claims. Finally, the removal of claim decomposition results in accuracy scores worse than the majority baseline. We observe that the removal of claim decomposition results in substantially more NEI predictions for longer claims, further discussed in Section \ref{sec:limitations}. %
\begin{table}[ht!]
\resizebox{1\linewidth}{!}{
	\begin{tabular}{ llll|cc} 
		\hline
            ArithExp & Constr. & Rationale & Decomp & Full & Num. \\ 
		\hline
            \cmark & \cmark & \cmark & \cmark & 72.0 & 71.4 \\
            \cmark & \xmark & \cmark & \cmark & 69.2 & 66.2 \\
            \xmark & \xmark & \cmark & \cmark & 66.1 & 61.0 \\
            \xmark & \xmark & \xmark & \cmark & 60.9 & 59.9 \\ 
            \cmark & \cmark & \cmark & \xmark & 66.3 & 63.7 \\
            \xmark & \xmark & \xmark & \xmark & 44.6 & 43.0 \\ 
            \hline
	\end{tabular}}
	\caption{Ablation study on the components of \projectname.%
    }
	\label{tab:ablation}
\end{table}

\paragraph{Qualiative comparison between \projectname and Binder.} We finally discuss some qualitative differences between \projectname and program-based approaches, such as Binder. We show an example of the entire output for both approaches (incl. claim decomposition) in Appendix Figures~\ref{app:end-to-end-example} and \ref{app:end-to-end-example2}. %
First, we observed greater flexibility with \projectname over Binder for reasoning when natural language is involved. Consider, for instance, small typographical errors, such as \emph{Asiacom Philippine}, instead of \emph{Asiacom Philippine}\underline{s}. This small typographical error results in different veracity predictions between \projectname and Binder, with the latter erroneously predicting the refuted label.
Binder's failure mode is partially caused by SQL's separation between query construction and table execution. This contrasts \projectname's framework of natural logic, where the content in the claim and evidence are directly compared.\footnote{It is possible to argue that Binder's issue is caused by incorrect parsing of the claim, as the SQL query itself should have accounted for the spelling error. However, the decision of equating \emph{Asiacom Philippine} with \emph{Asiacom Philippines} in the parse would consequentially be hidden from the user, while the natural logic proof explicitly states that it considers the two spellings of the entity equivalent.} 

Another issue with using SQL is the lack of a formal guarantee that all factoids in the claim are covered by the query. While Binder's SQL query in the same example incorporates that \emph{Asiacom Philippine} is \emph{a major stakeholder}, it does not state of whom (i.e.\ \emph{Globe Telecom}). Thus, if we were to find the same table for a different stakeholder, we would have derived the wrong verdict. In contrast, \projectname requires each part of the (sub-)claim to be incorporated in the proof. Finally, SQL is ill-suited for predicting \texttt{NEI} labels, as we also see from Binder's macro F$_1$ result in Table~\ref{tab:results-feverous}. Programs cannot inherently yield an \texttt{NEI} prediction, and instead, we have used programs which don't execute on a given table (i.e.\ output \texttt{null}) as a proxy. Yet, this signal is unreliable since it conflicts with the training objective of language models to produce executable queries. %

\section{Limitations}
\label{sec:limitations}

\begin{figure*}
            \centering
            \fbox{\begin{minipage}{38em}
                    \small    
                    \textbf{Claim:} In 2018, Ortegal had three municipalities and a population larger than 12,000. \\
                    
                    \hspace{18em}  \textbf{Subclaim 1 Proof:}\\
                    \vspace{0.3em}
                    \hspace{3em} \begin{tabular}{l | c c c}
                        \toprule
                         Claim Span $c_i$ & In 2018 &  Ortegal & had three municipalities \\
                         ArithExp $a_i$ & 2018 & COPY Ortegal & COUNT 4  \\
                         NatOp & $\equiv$ & $\equiv$ & $\downharpoonleft \! \upharpoonright$ \\
                         DFA State & \coloredcircle{celadon!50}{S} & \coloredcircle{celadon!50}{S} & \coloredcircle{babypink!50}{R} \\
                        \bottomrule
            	\end{tabular}
            
                    \vspace{0.75em}          
                    
                    \hspace{18em} \textbf{Subclaim 2 Proof:} \\
                    \vspace{0.3em}
                    \hspace{3em} \begin{tabular}{l | c c c}
                        \toprule
                          Claim Span $c_i$ & In 2018 &  Ortegal & had a population larger than 12,000 \\
                         ArithExp $a_i$ & 2018 & COPY Ortegal & SUM 12,238  \\
                         NatOp & $\equiv$ & $\equiv$ & $\sqsubseteq$\\
                         DFA State & \coloredcircle{celadon!50}{S} & \coloredcircle{celadon!50}{S} & \coloredcircle{celadon!50}{S} \\
                        \bottomrule
            	\end{tabular}
             
                    \vspace{0.75em}
                     \textbf{Verdict, original claim}: $\coloredcircle{celadon!50}{S} \xrightarrow{\equiv} \coloredcircle{celadon!50}{S} \xrightarrow{ \downharpoonleft \! \upharpoonright} \coloredcircle{babypink!50}{R} \xrightarrow{\sqsubseteq} \coloredcircle{paperblue!50}{N} \rightarrow $ \textbf{NEI.} \xmark\\
                    \textbf{Verdict, aggregation:} $\text{Aggr}(\coloredcircle{babypink!50}{S}, \coloredcircle{celadon!50}{R}) = \coloredcircle{babypink!50}{R} \rightarrow$ \textbf{Refutation.} \cmark
            \end{minipage}}
        \caption{Illustrating claim decomposition and verdict aggregation. Further, claim decomposition partially addresses the issue of producing less informative predictions by constraining natural logic to left-to-right execution. %
        }
        
        \label{fig:proof-decompositon-example}
    \end{figure*}

While the addition of arithmetic reasoning capabilities addresses a vital limitation of natural logic-based systems, \projectname \ is not attempting to modify natural logic's model of compositional entailment itself (i.e. the DFA in Figure \ref{fig:natlog-dfa}). 
Natural logic fails some inferences such as De Morgan's laws for quantifiers,  generally having less deductive power than first-order logic \citep{maccartney-manning-2014-natural, karttunen-2015-natural}. In contrast, \projectname \  incorporates relevant reasoning processes in the generated proof either explicitly, e.g.\ ArithExps, claim decomposition, or latently, e.g.\ the assignment of NatOps between an aligned claim and evidence span. %
This flexibility sometimes comes at the cost of
the proof’s granularity: Consider a natural-language instantiation of De Morgan's law from \citet{maccartney-2009-natural} where the claim ``\emph{Some birds do not fly}" is entailed by the evidence ``\emph{Not all birds fly}". Due to natural logic's limitations, the most fine-grained correct proof would be to align (\emph{Not all}, \emph{Some do not}), (\emph{birds}, \emph{birds}) and (fly, fly) to produce the proof $\coloredcircle{celadon!50}{S} \xrightarrow{\equiv} \coloredcircle{celadon!50}{S} \xrightarrow{\equiv} \coloredcircle{celadon!50}{S} \xrightarrow{\equiv}  \coloredcircle{celadon!50}{S}$; thus the reasoning between the negations and quantifiers in the aligned pair required to arrive at the set-theoretic relation is omitted from the proof itself. Therefore, the proofs of \projectname \ are not necessarily fully comprehensive explanations, as they do not fully explain the production of the proof.

Moreover, proofs of \projectname \ do not allow to assign NatOp sequences to individual claim spans, such as $\emph{cat} \rightarrow \emph{dog} \rightarrow \emph{poodle}$, which can be a limitation for multi-hop claims where multiple pieces of evidence from one or more tables have to be combined for a single span beyond arithmetic functions. Furthermore, proofs are produced and executed from left to right. However, natural logic does not impose such constraints and is instead non-deterministic by design. This can lead to inconsistencies, as the rearrangement of a NatOp sequence $O$ can lead to uninformative veracity predictions, i.e.\ NEI predictions \citep{maccartney-manning-2009-extended, angeli-etal-2016-combining}.
For instance, consider a variation of the running example shown in Figure \ref{fig:proof-decompositon-example}: ``\emph{In 2018, Ortegal had three municipalities and a population larger than 12,000.}". Assuming the same NatOp relations are assigned, an NEI verdict would be produced: \coloredcircle{celadon!50}{S} $\xrightarrow{\equiv} \coloredcircle{celadon!50}{S} \xrightarrow{ \downharpoonleft \! \upharpoonright} \coloredcircle{babypink!50}{R} \xrightarrow{\sqsubseteq}$  \coloredcircle{paperblue!50}{N}. \projectname \ mitigates this issue via two mechanisms: (i)  using claim decomposition to avoid long proofs where such phenomena occur, (ii) considering multiple proof candidates at different granularity levels, following \citet{aly-etal-2023-qanatver} (e.g.\ \emph{three municipalities and a population larger than 12,000} could be considered a single span with the $\downharpoonleft \! \upharpoonright$ NatOp). As shown in Figure \ref{fig:proof-decompositon-example}, by breaking the original claims into atomic units of information, the individual subclaim verdicts (via DFA transitions $\coloredcircle{celadon!50}{S} \xrightarrow{\equiv} \coloredcircle{celadon!50}{S} \xrightarrow{\equiv} \coloredcircle{celadon!50}{S} \xrightarrow{ \downharpoonleft \! \upharpoonright} \coloredcircle{babypink!50}{R}$ and $\coloredcircle{celadon!50}{S} \xrightarrow{\equiv} \coloredcircle{celadon!50}{S} \xrightarrow{\equiv} \coloredcircle{celadon!50}{S} \xrightarrow{\sqsubseteq} \coloredcircle{celadon!50}{S}$ for subclaim 1 and 2, respectively)
aggregate into the correct overall verdict. Both mechanisms also help in dealing with complex, multi-clause claims, where multiple erroneous and independent facts can lead to NEI predictions (double $\downharpoonleft \! \upharpoonright$)  -- another weak point of natural logic's nondeterministic composition of NatOps.

\section{Conclusion}
This paper presented \projectname, a natural logic inference system that adds arithmetic reasoning capabilities for few-shot fact verification on tables. We presented a set-theoretic definition between numerals in a claim and answers calculated on evidence via ArithExps. We proposed a method for leveraging LLMs to generate ArithExps via claim-aware question generation and rationale-guided question answering with constrained decoding. \projectname \ outperforms all baseline systems on FEVEROUS and in a domain-transfer setting on Tabfact, highlighting our model's generalizability. We show that \projectname \ %
has learned a nuanced understanding of numerals, more sensitive to the context of a claim than other baselines. Future work investigates natural logic for scalar implicature on diverse datasets, beyond Wikipedia, with different requirements for numerical precision. We further aim to explore the actionability of produced proofs in various decision-making processes when used as assistive tools for humans.

\section*{Acknowledgements}
This work was supported by the Engineering and Physical Sciences Research Council Doctoral Training Partnership (EPSRC). Andreas Vlachos is supported by the ERC grant AVeriTeC (GA 865958). The authors would like to thank Sana Kidwai for helpful conversations on linguistic concepts and Chenxi Whitehouse for useful discussions and feedback on the paper. We further thank the anonymous reviewers and the action editor Kenji Sagae for their valuable and detailed feedback. 

\newpage

\bibliography{zotero-dissertation}
\bibliographystyle{acl_natbib}

\clearpage

\newpage

\appendix

\section{Method Details}
\label{app:method-details}
Table \ref{tab:natops} shows the set-theoretic definitions of NatOps $o \in \mathbb{O}$. The effect of environments on the entailment relations is modelled in natural logic via projection functions $\rho: \mathbb{O} \rightarrow \mathbb{O}$ \citep{maccartney-manning-2009-extended}. The upward-entailing environment is the default environment with the projection function being the identity. Table \ref{tab:tabver-projection-function} shows the projection function $\rho_{\downarrow}$ for downward-monotone environments. It further shows the projection function $\rho_{\text{num}\downarrow}$  for a number's reading in a downward-monotone environment that results from such a lower-bounded reading of numerals. The prompt templates for $M_{QG}$ and $M_{QA}$ are shown in Listing \ref{qg_prompt} and \ref{qa_prompt}, respectively. 

\begin{table}[ht!]
\centering
	\footnotesize
	\begin{tabular}{ |l|c|} 
		\hline
		NatOp & Set-theoretic  \\
		\hline
		Equivalence ($\equiv$)  & $x = y$ \\
		Frw. Entailment ($\sqsubseteq$) & $x \subset y $ \\
		Rev. Entailment ($\sqsupseteq$) & $ x \supset y $\\
		Negation ($\curlywedge$) & $x \cap y  = \emptyset \land  x \cup y = U$   \\
		Alternation ($\downharpoonleft \! \upharpoonright$) & $ x \cap y = \emptyset \land  x \cup y \neq U $  \\
		Independence (\#) & All other cases   \\
		\hline
	\end{tabular}
	\caption{Natural logic operators (NatOps) and their set-theoretic definitions.}
	\label{tab:natops}
\end{table}

\begin{table}[ht]
    \centering
    \begin{tabular}{c | c c c c c c}
          o &  $\mynatop{equiv}$ &  $\mynatop{forwardentailment}$ & $\mynatop{reventailment}$ & $\mynatop{alternation}$ & $\mynatop{negation}$ & $\mynatop{independence}$ \\
         \hline
           $\rho_{\uparrow}(o)$ &  $\mynatop{equiv}$ &  $\mynatop{forwardentailment}$ & $\mynatop{reventailment}$ & $\mynatop{alternation}$ & $\mynatop{negation}$ & $\mynatop{independence}$ \\
         $\rho_{\downarrow}(o)$ &  $\mynatop{equiv}$ &  $\mynatop{reventailment}$ & $\mynatop{forwardentailment}$ & $\mynatop{cover}$ & $\mynatop{negation}$ & $\mynatop{independence}$ \\
         $\rho_{\text{exactly one}}(o)$ &  $\mynatop{equiv}$ &  $\mynatop{independence}$ & $\mynatop{independence}$ & $\mynatop{independence}$ & $\mynatop{independence}$ & $\mynatop{independence}$ \\
         \hline
         $\rho_{\text{num}\uparrow}(o)$ & $\mynatop{equiv}$ &  $\mynatop{forwardentailment}$ & $\mynatop{reventailment}$ & $\mynatop{alternation}$ & $\mynatop{negation}$ & $\mynatop{independence}$ \\
         $\rho_{\text{num}\downarrow}(o)$ &  $\mynatop{forwardentailment}$ &  $\mynatop{reventailment}$ & $\mynatop{forwardentailment}$ & $ \mynatop{forwardentailment}, \mynatop{reventailment}$ & $\mynatop{cover}$ & $\mynatop{independence}$\\
    \end{tabular}
    \caption[Projection functions when including numbers.]{Top shows the general projection functions for upward- and downward-monotone environments, as well as for the quantifier \emph{exactly one}. Bottom shows how these projection functions specifically affect the reading of numerals in an exact reading ($\rho_{\text{num}\uparrow}$) and an at least reading ($\rho_{\text{num}\downarrow}$). The cover ($\mynatop{cover}$) NatOp is shown here for completeness only. %
    }
    \label{tab:tabver-projection-function}
\end{table}

\begin{figure*}[!t]
\begin{lstlisting}[caption={The prompt template for the question generation model $M_{QG}$}, label={qg_prompt}]
"system_description": "Always ensure that each generated question is a single sentence. Ensure to enumerate each question and corresponding answer with the same number.",
"task_description": "For a given claim, generate questions with answers that are found in the claim itself. Ensure that the answer is located exactly as a substring in the claim itself. Generate the least number of questions as needed to verify the key information of the claim.",
"task_description_arithmetic":"For a given claim, generate questions with answers that are found in the claim itself. Only generate questions that require arithmetic reasoning, such as counting, addition, or averaging. Ensure that the answer is located exactly as a substring in the claim itself. Generate the least number of questions as needed to verify the key information of the claim.",
"demo_sep": "\n\n\n",
"context_format": "{INST}\n\nClaim: {}\nQuestions: {}",
"response_template": "\nQuestions:",
\end{lstlisting}
\end{figure*}
\begin{figure*}[!t]
\begin{lstlisting}[caption={The prompt template for the question answering model $M_{QA}$}, label={qa_prompt}]
    "system_description": "Ensure that the answer to each question starts on a new line, separated by a single newline character.",
    "task_description": "Read the tables below taken from Wikipedia to answer a question regarding the table. Use only the following functions for arithmetic reasoning and state them explicitly when used: {FUNCTIONS}. If none of these functions is needed for reasoning say N/A.",
    "demo_sep": "\n\n\n",
    "context_format": "{INST}\nQuestion: {}\nTable: {}\nAnswer: {}",
    "answer_format": "Extracted evidence from table: {EVIDENCE} Computation: {COMPUTATION} Result: {ARITHEXP}",
    "response_template": "\nAnswer:",
    "negative_answer": "N/A",
    "negative_evidence": "Not sufficient information found in the table.",
\end{lstlisting}
\end{figure*}

\paragraph{Training \& Hyperparameters}  We fine-tune the question generation $M_{QG}$ and question answering $M_{QA}$ model using default hyperparameters. Specifically, we use a learning rate of $2^{-4}$ and train for a total of $10$ epochs across all models and experiments. The maximum generation length for $M_{QG}$ is set to $100$ tokens for the generation of question, and the constraint answer selection is set to any-length span in $c$. For $M_{QA}$ the maximum length of $E_{rel}$ is 100 tokens between every generated number. We use adamw \citep{loshchilov-hutter-2018-decoupled} as the optimizer. We use a batch size of $1$ during training with gradient accumulation, resulting in an effective batch size of $8$. For LoRA, we use a rank $r=16$ and apply it to the query and value vectors of the attention mechanism. For fine-tuning, we exclude tokens of the prompts from the loss computation that are not part of the gold answer, so we are not fine-tuning the instructions, only the answers that follow after the instruction. For our proof generation model $M_{P}$ we use the default hyperparameters of QA-NatVer \citep{aly-etal-2023-qanatver}.

\section{Dataset Details}
\label{app:dataset-details}

Quantitative characteristics of the tabular subset of FEVEROUS are shown in Table \ref{tab:feverous-tables-stats}. %
The table further shows the statistics for the function annotations of $160$ claims. The claims were sampled randomly and annotated by the authors of the paper as function annotations are made irrespectively of any model, limiting potential biases. Note that annotations are in a multi-label format since multiple functions can be required to verify a single claim. %
\begin{table}[ht!]
\resizebox{1\linewidth}{!}{
	\begin{tabular}{ |c|c|c|} 
		\hline
		Property & All & Numerical Subset \\
		\hline
		Number of claims  & 2011 &  521 \\
		Claims with more than 1 table & 129 & 36 \\
		Supported claims & 704 (35\%) & 178 (34.1\%) \\
		Refuted claims & 1242 (61.7\%) & 338 (64.8 \%)\\
		NEI claims & 65 (3.2\%) & 5 (1\%) \\
		\hline
		Avg. number of rows & 14.3 & 15.1 \\
		Avg. number of col & 4.82 & 5.9 \\
		Avg. num highlighted cells & 4.85 & 7.3 \\
		\hline
		\multicolumn{3}{|c|}{Function annotations on 160 samples} \\
		\hline
		Num. COPY &  143 & -- \\
  		Num. COMPARATIVES & 12 & -- \\
		Num. COUNT & 28 & --  \\
		Num. SUM & 1 & -- \\
            Num. DIFF & 6 & -- \\
  		Num. MIN/MAX & 3 & -- \\
		\hline	
	\end{tabular}}
	\caption{Quantitative characteristics of the tabular FEVEROUS subset.}
	\label{tab:feverous-tables-stats}
\end{table}

\section{Implementation Details}
\label{app:impl-details}
\begin{figure*}[!t]
\begin{lstlisting}[caption={The prompt template for the claim decomposition model $M_{D}$}, label={decomp_prompt}]
    "system_description": "Ensure to enumerate each subclaim and that each subclaim starts on a new line.",
    "task_description": "Break a claim down into its subclaims. Ensure that each subclaim can stand by itself and does not require the original claim to be understood. Always ensure that each generated subclaim is a single sentence without conjunctive clauses.",
    "demo_sep": "\n\n\n",
    "context_format": "{INST}\n\nClaim: {}\nSubclaims: {}",
    "response_template": "\nSubclaims:",
\end{lstlisting}
\end{figure*}

\projectname was implemented using PyTorch \citep{paszke-etal-2019-pytorch}, making use of PyTorch Lightning \citep{wa-2019-pytorch}, and Huggingface's model checkpoints.  We use the Huggingface's SFTTrainer\footnote{\url{https://huggingface.co/docs/trl/en/sft_trainer}} to fine-tune $M_{QG}$, $M_{QA}$, and $M_{D}$. The prompt template for the decomposition model $M_{D}$ is shown in Listing \ref{decomp_prompt}. We use the Huggingface checkpoints for LLama2-7B\footnote{\url{https://huggingface.co/meta-llama/Llama-2-7b-chat-hf}}, MistralOrca-7B\footnote{\url{https://huggingface.co/Open-Orca/Mistral-7B-OpenOrca}}, TAPAS\footnote{\url{https://huggingface.co/google/tapas-large}}, and TAPEX\footnote{\url{https://huggingface.co/microsoft/tapex-large}}. The PASTA checkpoint is taken from the associated repository\footnote{\url{https://github.com/ruc-datalab/PASTA}}. For constrained decoding, we used the library guidance-ai\footnote{\url{https://github.com/guidance-ai}}. The Mistral models are licensed under Apache2.0 and Llama2 is licensed under the llama license\footnote{\url{https://github.com/facebookresearch/llama/blob/main/LICENSE}}. Our research is consistent with the licenses' intended use. The models are intended for use in English. All experiments are run on a single Quadro 8000 with 48GB memory. To fine-tune $M_{P}$ with a Flan-T5-3B backbone we use a single A100 80GB.

\paragraph{Baselines}
We use the available implementations for LPA\footnote{\url{https://github.com/wenhuchen/Table-Fact-Checking}}, SASP\footnote{\url{https://github.com/ousuixin/SASP}}, and Binder\footnote{\url{https://github.com/xlang-ai/Binder}}.
Following their original implementation, all three neuro-symbolic baselines consider the first table row of a table as the header row, both for TabFact and FEVEROUS. Further, Binder generates its SQL query conditioned only on the first three rows of a table. LPA was trained for 20 epochs since the default number of training epochs (10) was not sufficient to reach convergence. Binder uses the default hyperparameters as specified for Tabfact, but we use $4$ instead of $18$ in-context examples as MistralOrca, with a full number of in-context examples, produced empty answers very frequently. We hypothesise that MistralOrca generates the end of text token too early since the OpenOrca dataset, on which MistralOrca-7B has been instruction-tuned, consists of gold answers which are in 93\% of the cases shorter than 2.5K tokens.%
We train DeBERTa, PASTA, TAPEX, and TAPAS using the HuggingFace Trainer for 10 (100) epochs with full (few-shot) data and a learning rate of $1 \times 10^{-5}$.\footnote{\url{https://huggingface.co/docs/transformers/en/main_classes/trainer}} To sanity check our training pipeline, we trained TAPAS in a full supervision setting on TabFact's 92,283 training instances, achieving a score of 82.1 accuracy points versus 81.59 points via the official model checkpoint\footnote{\url{https://huggingface.co/google/tapas-large-finetuned-tabfact}}. To fine-tune the LLM baselines with LoRA, we use Huggingface's SFTTrainer\footnote{\url{https://huggingface.co/docs/trl/en/sft_trainer}}.

\section{Reading of numerals probe - Details}
\label{app:reading-numeral-details}

We construct the probing dataset by first filtering instances from the FEVEROUS evaluation data labeled as Supported that contain numbers, excluding dates (e.g. 1939) but including percentages and floating point numbers. Afer a further manual inspection a total $91$ claims remain. For each claim, we generate $17$ variations of a numeral $x$:

{\noindent
\footnotesize
\begin{enumerate*}[label=(\arabic*),itemjoin=\\]
\item $x+1$ (Adding one)
\item $x + x * 0.02$ (Adding 2\%)
\item $x - x * 0.02$ (Subtracting 2\%)
\item $x + x * 0.1$ (Adding 10\%)
\item $x - x * 0.1$ (Subtracting 10\%)
\item $x + x * 0.25$ (Adding 25\%)
\item $x - x * 0.25$ (Subtracting 25\%)
\item rounding via closest number to $x$ that satisfies: $\frac{x'}{1 * 10^y} \leq 9$ (10-ness)
\item rounding via closest number to $x$ that satisfies: $\frac{x'}{5 * 10^y} \leq 9$ (5-ness)
\item rounding via closest number to $x$ that satisfies: $\frac{x'}{2.5 * 10^y} \leq 9$ (2.5-ness)
\item (About$\mid$Around$\mid$Approximately) + 10-ness (Modifier 10-ness)
\item (About$\mid$Around$\mid$Approximately) + 5-ness (Modifier 5-ness)
\item (About$\mid$Around$\mid$Approximately) + 2.5-ness (Modifier 2.5-ness)
\item 'At most' + (Subtracting 10\%) (Cardinal at most, incorrect)
\item 'At least' + (Adding 10\%) (Cardinal at least, incorrect)
\item 'At most' + (Adding 10\%) (Cardinal at most, correct)
\item 'At least' + (Subtracting 10\%) (Cardinal at least, correct)
\end{enumerate*}}

Rounding to numbers that satisfy the 10-ness, 5-ness, and 2.5-ness property follows the empirical observation by \citet{jansen-pollmann-2001-numbers} that round numbers satisfying this arithmetic property occur more frequently than round numbers that do not. For instance, the number $1010$ does not satisfy either 10-ness, 5-ness, or 2.5-ness and would generally be considered an atypical way of rounding ($1000$ would most likely be more natural). We follow the terminology of \citep{keenan-2017-quantifier} to describe the modifiers \emph{at most} and \emph{at least} as cardinal determiners. %

We categorize these variations into the numerical classes shown in Table \ref{tab:pragmatic halo} as follows:
{
\footnotesize

Inaccuracy $\Delta$ +1: Variation 1.

Inaccuracy $\Delta$ 2\%: Average of Variation 2 + 3.

Inaccuracy $\Delta$ 10\%: Average of Variation 4 + 5.

Inaccuracy $\Delta$ 25\%: Average of Variation 6 + 7.

Rounding: Average of Variation 8 + 9 + 10.

Modifiers: Average of Variation 11 + 12 + 13.

Cardinal (incorrect): Average of Variation 14 + 15.

Cardinal (correct): Average of Variation 16 + 17.
}

\clearpage

 \begin{figure*}[ht!]
            \centering
            \fbox{\begin{minipage}{40em}
            \textbf{Claim}: John McCain won 16 delegates during the 2008 Washington Republican presidential primary, which is eight more than Mike Huckabee won.
            
            \textbf{Evidence}: 2008 United States presidential election in Washington (state). Caucuses Republican caucuses and primary
            \begin{tabular}{|r r r r r|}
		\toprule
		\textbf{Candidate} & \textbf{State delegate} & \textbf{Percentage}& \textbf{Delegates} & \textbf{Counties carried} \\ 
		John McCain & 3,228 & 25.9\% & 	16 &	11 \\
            Mike Huckabee	&2,959&	23.52\%	& 8 &	11 \\
            ... & ... & .. & .. & ... \\
		\bottomrule
    	\end{tabular}

            \textbf{Generated Subclaim 1:} John McCain won 16 delegates during the 2008 Washington Republication presidential primary.

            \textbf{Generated Subclaim 2:} Mike Huckabee won eight fewer delegates than John McCain.
            
            \noindent\rule{40em}{0.4pt}
            \textbf{Binder} (SQL): 

            Subclaim 1: SELECT ((SELECT SUM(delegates) FROM w WHERE candidate = "john mccain") = 16\\
            \textbf{Predicted Verdict}: SUPPORTS
            
            Subclaim 2: SELECT ((SELECT SUM(delegates) FROM w WHERE candidate = "mike huckabee") - (SELECT SUM(delegates) FROM w WHERE candidate = "john mccain")) = -8 \\
            \textbf{Predicted Verdict}: SUPPORTS
            \textbf{Overall Verdict}: SUPPORTS
            
            \noindent\rule{40em}{0.4pt}
            \vspace{1em}
            \textbf{\projectname}:
            Subclaim 1: \\
            \textbf{Generated Questions}: 1. How many delegates did John McCain win during the 2008 Washington Republican presidential primary? 1. Associated claim Span ($c_i$): 16

            \textbf{Question Answering}: John McCain won 16 delegates in the 2008 Washington Republican presidential primary. Computation: No computation is required. ArithExp ($a_i$): COPY 16

            \textbf{Generated Proof}:
            
            \begin{tabular}{|r r r r|}
		\toprule
		\textbf{Claim Span} & \textbf{Evidence Span} & \textbf{NatOP} & \textbf{DFA State} \\ 
		John McCain & John McCain & = & \coloredcircle{celadon!50}{S} \\
            won & won & = & \coloredcircle{celadon!50}{S} \\
            16 delegates & COPY 16 & = & \coloredcircle{celadon!50}{S} \\
            during the 2008 Washington &  in the 2008 Washington & & \\
            Republican presidential primary & Republican presidential primary & = & \coloredcircle{celadon!50}{S}\\
		\bottomrule
    	\end{tabular}

            \textbf{Predicted verdict:} SUPPORTS

            \vspace{1em}
     
            Subclaim 2: \\
            \textbf{Generated Questions}: 1. How many fewer delegates did Mike Huckabee win than John McCain at the 2008 Washington Republican presidential primary? 1. Associated claim span ($c_i$): eight

            \textbf{Question Answering}: Mike Huckabee won 8 delegates and John McCain won 16 delegates. Computation: 8 - 16 = -8. ArithExp($a_i$): COMP -8

            \begin{tabular}{|r r r r|}
		\toprule
		\textbf{Claim Span} & \textbf{Evidence Span} & \textbf{NatOP} & \textbf{DFA} \\ 
		Mike Huckabee won  & & \\
            eight fewer delegates than John McCain & COMP -8 & = & \coloredcircle{celadon!50}{S} \\
            at the 2008 Washington &  in 2008 Washington & &  \\
            Republican presidential primary & Republican presidential primary & = & \coloredcircle{celadon!50}{S}\\
		\bottomrule
    	\end{tabular}

            \textbf{Predicted verdict:} SUPPORTS  \textbf{Overall verdict:} SUPPORTS

            \end{minipage}}
        \caption{Output produced by \projectname. For comparison, we show Binder's output. While more compact, Binder's SQL query is less readable and omits relevant context from its query (2008 Washington primary). }
        \label{app:end-to-end-example}
\end{figure*}

\clearpage

 \begin{figure*}[ht!]
            \centering
            \fbox{\begin{minipage}{40em}
            \textbf{Claim}: Asiacom Philippine, Inc. stands as the major shareholder of Globe Telecom, with 50.85\% of total shares.
            
            \textbf{Evidence}: Globe Telecom. Ownership.\\
            \begin{tabular}{|r r r r|}
		\toprule
        	\textbf{Major Shareholder} &	\textbf{\% of Total}	 & \textbf{Common Shares}	& \textbf{Preferred Shares}\\
                Ayala Corporation & 	13.20\% & 	41,157,276 &	— \\
                SingTel Group &	20.10\% &	62,646,487 &	— \\
                Asiacom Philippines, Inc.	& 50.85\%	& —	&158,515,016 \\
		\bottomrule
    	\end{tabular}

            \textbf{Generated Subclaim 1:} 
             Asiacom Philippine, Inc. stands as the major shareholder of Global Telecom.

            \textbf{Generated Subclaim 2:} Asiacom Philippine, Inc. has 50.85\% of total shares.
            
            \noindent\rule{40em}{0.4pt}
            \textbf{Binder} (SQL): 

            Subclaim 1:  SELECT (SELECT major shareholder FROM w WHERE row\_id = 2) = "Asiacom Philippine, Inc.") \\
            \textbf{Predicted Verdict}: REFUTES
            
            Subclaim 2: SELECT (SELECT major shareholder, \% of total FROM w WHERE row\_id = 2) = ("Asiacom Philippine, Inc.", "50.85\%") \\
            \textbf{Predicted Verdict}: REFUTES
            \textbf{Overall Verdict}: REFUTES 
            
            \noindent\rule{40em}{0.4pt}
            \vspace{1em}
            \textbf{\projectname}:

            Subclaim 1: \\
            \textbf{Generated Questions}: 1. Who is the major shareholder of Globe Telecom? 1. Associated claim Span ($c_i$): Asiacom Philippine, Inc

            \textbf{Question Answering}: Ayala Corporation, SingTel Group, and Asiacom Philippines, Inc. are major shareholders of Globe Telecom. Computation: No computation is required. ArithExp ($a_i$): FILTER Ayala Corporation, SingTel Group, and Asiacom Philippines, Inc.

            \textbf{Generated Proof}:
            
            \begin{tabular}{|r r r r|}
		\toprule
		\textbf{Claim Span} & \textbf{Evidence Span} & \textbf{NatOP} & \textbf{DFA} \\ 
		Asiacom Philippine, Inc & FILTER Ayala Corporation, SingTel Group & & \\
            stands as the major shareholder & , and Asiacom Philippines, Inc. & $<$ & \coloredcircle{celadon!50}{S} \\
            of Globe Telecom & of Globe Telecom & = & \coloredcircle{celadon!50}{S} \\
		\bottomrule
    	\end{tabular}

            \textbf{Predicted verdict:} SUPPORTS

            \vspace{1em}
     
            Subclaim 2: \\
            \textbf{Generated Questions}: 1. What percentage of total shares does Asiacom Philippine, Inc. have? 1. Associated claim span ($c_i$): 50.85\%

            \textbf{Question Answering}: Asiacom Philippines, Inc. has 50.85\% of total shares. No computation is required. ArithExp($a_i$): FILTER 50.85\%

            \begin{tabular}{|r r r r|}
		\toprule
		\textbf{Claim Span} & \textbf{Evidence Span} & \textbf{NatOP} & \textbf{DFA state} \\ 
		Asiacom Philippine, Inc.   & Asiacom Philippines, Inc. & = & \coloredcircle{celadon!50}{S}\\
            has 50.85\% & FILTER 50.85\% & = & \coloredcircle{celadon!50}{S} \\
            of total shares &  of total shares & = & \coloredcircle{celadon!50}{S}\\
		\bottomrule
    	\end{tabular}

            \textbf{Predicted verdict:} SUPPORTS

            \vspace{1em}

            \textbf{Overall verdict:} SUPPORTS

            \end{minipage}}
        \caption{Output produced by \projectname. For comparison, we show Binder's output. In contrast to \projectname \ in its natural logic formulation, Binder's query is unable to handle probabilistic enrichment of natural language, for instance, due to typing errors (e.g. Asiacom Philippine\textbf{s}, Inc.).}
        \label{app:end-to-end-example2}
\end{figure*}


\end{document}